# Applications of Generative Adversarial Networks in Neuroimaging and Clinical Neuroscience

Rongguang Wang,[1,2] Vishnu Bashyam,[1,2] Zhijian Yang,[1,2] Fanyang Yu,[1,2] Vasiliki Tassopoulou,[1,2] Sai Spandana Chintapalli,[1,2] Ioanna Skampardoni,[1,2,3] Lasya P. Sreepada,[1,2] Dushyant Sahoo,[1,2] Konstantina Nikita,[3] Ahmed Abdulkadir,[1,2,4] Junhao Wen[1,2] and Christos Davatzikos[1,2,5]

## Abstract

Generative adversarial networks (GANs) are one powerful type of deep learning models that have been successfully utilized in numerous fields. They belong to the broader family of generative methods, which learn to generate realistic data with a probabilistic model by learning distributions from real samples. In the clinical context, GANs have shown enhanced capabilities in capturing spatially complex, nonlinear, and potentially subtle disease effects compared to traditional generative methods. This review critically appraises the existing literature on the applications of GANs in imaging studies of various neurological conditions, including Alzheimer's disease, brain tumors, brain aging, and multiple sclerosis. We provide an intuitive explanation of various GAN methods for each application and further discuss the main challenges, open questions, and promising future directions of leveraging GANs in neuroimaging. We aim to bridge the gap between advanced deep learning methods and neurology research by highlighting how GANs can be leveraged to support clinical decision making and contribute to a better understanding of the structural and functional patterns of brain diseases.

**Author affiliations:**

1 Center for AI and Data Science for Integrated Diagnostics, University of Pennsylvania, Philadelphia, USA

2 Center for Biomedical Image Computing and Analytics, University of Pennsylvania, Philadelphia, USA

3 School of Electrical and Computer Engineering, National Technical University of Athens, Athens, Greece

4 Department of Clinical Neurosciences, Lausanne University Hospital and University of Lausanne, Lausanne, Switzerland

5 Department of Radiology, Perelman School of Medicine, University of Pennsylvania, Philadelphia, USA

Corresponding authors: Rongguang Wang, Christos Davatzikos

3700 Hamilton Walk, 7th Floor, Center for Biomedical Image Computing and Analytics, University of Pennsylvania, Philadelphia, PA 19104, USA

E-mail: rgw@seas.upenn.edu, christos.davatzikos@pennmedicine.upenn.edu

# Introduction

Advances in medical imaging techniques, including magnetic resonance imaging (MRI) and positron emission tomography (PET), have provided *in vivo* imaging-derived phenotypes capturing patterns of brain development, aging as well as of various diseases and disorders.[1-3] Embracing the "big data" era, the medical imaging community has widely adopted artificial intelligence (AI) for data analysis, from traditional statistical methods to machine learning (ML) models, which provides promise toward clinical translation.[4-6] Statistical tools such as univariate and multivariate prediction models are empowered to learn the associations between structural/functional variability and cognitive/psychiatric symptomatology in the human brain. Notably, advanced AI techniques have been successfully utilized in numerous clinical applications, such as computer-aided diagnosis, disease biomarker identification, and personalized disease risk quantification, which are bound to further revolutionize medical research and clinical practice. Among these techniques, deep learning (DL) has drawn increasing attention in medical imaging. DL algorithms are powerful in capturing the complex non-linear relationships between input features, thereby extracting low-to-high level latent features that are predictive of the response of interest.[7-11] So far, DL has been widely adopted in medical image processing tasks such as registration, reconstruction, segmentation, and synthesis, and analysis tasks such as disease diagnosis, anomaly detection, and pathology and prognosis evolution prediction.

Generative adversarial networks (GANs), first introduced in 2014 by Goodfellow et al.,[12] have had a profound influence in DL, leading to numerous applications. GAN is a generative method which synthesizes realistic-looking features/images by learning the sample distribution from real data. GAN and its variants have shown great promise in image generation tasks such as image enhancement, cross-modality synthesis, text-to-image synthesis and image-to-image translation.[13,14] This technique is particularly promising for neuroimaging and clinical neuroscience applications because it is capable of discovering and reproducing the complex and non-linear pathology patterns from medical images and data.

Many previous reviews of GANs focused on technical details of image synthesis in medical imaging.[15-19] However, less attention has been paid to the adoption of GANs in clinical neuroimaging studies. In this review, we present the current state of GANs in neuroimaging research, in various applications including neurodegenerative disease diagnosis, cancer and anomaly detection, brain development modeling, dementia trajectory tracing, lesion evolution prediction, and tumor growth estimation. By showcasing these different applications discussed in the literature, we demonstrate the advantages of GANs for neuroimaging studies, compared to traditional ML methods. We also discuss the current limitations of GANs and potential opportunities for adopting GANs in future neuroimaging research.

**Review perspective.** Our review emphasizes discovery and analysis of imaging phenotypes associated with neurological diseases via deep learning techniques, focusing on GANs in particular. We exclude papers solely focusing on methodology development for tasks such as image synthesis, registration, segmentation, reconstruction, modality translation, and dataset enlargement. We filter research articles on Google Scholar, PubMed, and several pre-print platforms containing words such as 'GAN', 'Generative Adversarial Network', 'Medical Imaging', and 'Brain'. We further screen the titles and abstracts for a thematic match. To provide a thorough review, we build connection graphs for each included paper using Semantic Scholar in order to find additional relevant publications. Finally, detailed examination of the methods and results of each paper helped us decide if they fall in our review scope. Based on the application areas of the selected papers, we divide them into two main categories: clinical diagnosis and disease progression. Each category is further split into finer tasks, which are described in the manuscript organization section.

**Manuscript organization.** The rest of the paper is organized as follows. In Sec. 2, we provide background knowledge on GANs and their algorithmic extensions for applications in neuroimaging. From Sec. 3 to Sec. 4, we comprehensively illustrate the applications of GANs in neurological research using imaging phenotypes. We discuss clinical diagnosis, including disease classification, with a primary focus on Alzheimer's disease and brain tumor detection in Sec. 3. In Sec. 4, we present modeling of imaging patterns of brain change in cognitively unimpaired brain aging and in several diseases, including Alzheimer's disease, brain lesion evolution and tumor growth. For each application, we introduce the background and challenges, describe the essential methodology for tackling the problem, showcase its advantages and promises from evaluation results, as well as critique the limitations and pitfalls. Finally, in Sec. 5, we suggest potential promising future directions and discuss open questions for each neurological application utilizing GANs, based on current issues and challenges.

# Preliminaries on GANs

GANs have the ability to approximate complex probability distributions and thereby generate realistic patterns or images, as well as capture effects of pathologic processes on imaging phenotypes. We will describe the mechanism of the standard GAN and its usage in neuroimaging studies (see Table 1). Then, we showcase a few GAN variants whose architectures have been modified to suit specific clinical tasks, such as disease diagnosis and prognosis.

## Original GAN

GAN was first proposed by Goodfellow et al.[12] to overcome the intractable probabilistic computation difficulty that deep generative models, such as the deep Boltzmann machine,[20] usually suffer from. There are two components in a GAN: the generator and the discriminator, as shown in Fig. 1A. Intuitively, we can think of the framework mechanism as a two-player game – player A and player B competing with each other to produce fake images and detect them. The game drives both parties to improve their techniques until the fake images are indistinguishable from the real images. Given a finite collection of data points $x$ sampled from the natural distribution, which is unknown, we would like to learn or approximate the natural distribution from the observations. A generator is defined to be a mapping function that projects noise variables sampled from a prior distribution to the data space. The prior distribution can be uniform or Gaussian, and the generator is parametrized by differentiable neural networks. Given the output of the generator and the real observations, the discriminator, which is also parameterized by neural networks, outputs the probability that the input comes from a real sample distribution. In this two-player game, we simultaneously train the generator to minimize the probability that the discriminator treats the generated image as fake and train the discriminator to maximize the probability of identifying the generated image as fake. This technique is called adversarial training.

At convergence, the discriminator should theoretically output 50% probability for any input and the generator produces samples that are indistinguishable from the real data. One advantage of GAN is that it can generate clear and high quality images whereas another popular deep generative model, the variational autoencoder (VAE),[21] can only produce blurry figures. Thus, GANs are well-suited for many applications in neuroimaging research, such as generating heterogeneous pathological patterns by mapping a healthy control image to potential reproducible disease signatures for subtype discovery. GANs can also predict the evolution of brain lesions or tumors for personalized disease diagnosis and prognosis. We discuss several popular variants of GANs that have been adapted for different areas of neuroimaging research in the next subsection.

# Variants of GAN

Based on the original GAN model, different variants were proposed in recent years for two main purposes: solving limitations of the original GAN and adapting it to different applications. Here, we introduce the main variants that have been applied in the neuroimaging studies that will be discussed in the following sections.

### Challenge oriented variants

Though the original GAN model has shown promising performance in generating realistic high-dimensional data, it still suffers from problems such as unstable optimization during training, mode collapse (learning to generate images following distributions of only a subset of training images), and poor quality (visually chaotic or blurry images) of generated data. Many GAN variants were proposed for solving these issues and also proved to be helpful in generating high-quality neuroimaging data.[22-24]

**Wasserstein GAN (W-GAN).** W-GAN[25] is one of the important variants proposed to address unstable optimization and mode collapse. Compared to the original GAN model, W-GAN shares a similar min-max training procedure but has a different loss function. With the new loss function, the training procedure aims to minimize the Wasserstein distance between distributions of generated data and real data, which is shown to be a better distance measure for image synthesis problems. The WGAN-GP[26] (W-GAN with gradient penalty) model was introduced as one improvement on W-GAN for more stable training with the gradient penalty method.

Besides variants in loss function, some other works propose to modify model structures for improving the quality of generated images.

**Deep convolutional GAN (DC-GAN) and progressive growing GAN (PG-GAN).** DC-GAN[27] is one of the earliest models that uses convolutional layers in both generator and discriminator for stable generation of higher quality RGB images. PG-GAN[28] (Fig. 1B), further achieves large high-resolution image generation by progressively increasing the number of layers during the training process. Systematic addition of the layers in both generator and discriminator enables the model to effectively learn from coarse-level details to finer details.

**Self-attention GAN (SA-GAN).** SA-GAN[29] leverages a self-attention mechanism in convolutional GANs. The self-attention module, complementary to convolutions, helps with modeling long range, multi-level dependencies across image regions, and thus avoids using only spatially local properties for generating high-resolution images.

## Application oriented variants

Besides addressing broader methodological challenges above, many other variants of GANs were developed for specific applications. Several applications in computer vision are also of great interest to the neuroimaging community, including informative latent space and conditional image generation.

**Informative latent space**

The latent vector in the GAN model is conventionally used as a random input for generating images, but does not have clear correspondence with the generated output data in an interpretable way. An informative latent space will help people interpret both the generative model and generated data, and make better use of them. In the field of neuroimaging, an informative latent space can be a low-dimensional representation for uncovering disease related imaging patterns.[22,30] Therefore, there are several variants proposed along this direction to make the latent vector correspond to features of generated images in an easily interpretable way:

**Info-GAN and Cluster-GAN.** Info-GAN[31] (Fig. 1E) divides latent variable into two parts $z$ and $c$ and enables $c$ to explicitly explain features in generated data $G(z, c)$. The model introduces a parameterized approximation of inverse posterior distribution $Q(c|G(z, c))$ which helps maximize the mutual information between latent variables and generated data, $I(c, G(z, c))$. Minimization of mutual information through $Q(c|G(z, c))$ can also be understood as a regularization on inverse reconstruction of latent variables from generated data. Therefore, compared with basic GAN, Info-GAN alternatively solves an information-regularized minimax game. Similar to Info-GAN, Cluster-GAN[32] shared the idea of reconstructing latent variables from generated data, but employed discrete latent variables, further enabling clustering through the latent space.

**Conditional image generation**

Conditional image generation, including image-to-image translation, generates images using some prior information instead of random input. For example, in the field of neuroimaging, there are several works approaches focusing on transformation of neuroimages among different modalities[33-35] and generation of neuroimages based on clinical information.[36] Most of these works are related to the following variants:

**Conditional-GAN (C-GAN).** C-GAN[37] (Fig. 1C) allows extra information, $y$, to be fed to both generator and discriminator, and thus is able to generate data x based on $y$ information. $y$ can be specific clinical information or the corresponding image in the source modality for image-to-image translation tasks.

**Cycle-GAN.** Paired data $y$ are not available in many cases, especially in the neuroimaging field. Dealing with this problem, Cycle-GAN[38] (Fig. 1D) enables unpaired image-to-image translation by imposing a specific cycle consistency loss for regularization besides standard GAN loss. The model has two mapping functions, $G$ and $F$, which transform data from source to target and from target to source domain respectively, while encouraging that generated output data can be reconstructed back to the input data, i.e. $F(G(x)) \approx x$ and $G(F(y)) \approx y$.

**Reversible-GAN (Rev-GAN).** Reversible-GAN[39] is an extension of Cycle-GAN. By utilizing invertible neural networks, the model possesses cycle-consistencies by design without explicitly constructing an inverse mapping function, thus achieving both output fidelity and memory efficiency.

**Multimodal unsupervised image-to-image translation GAN (MUNIT-GAN).** Both Cycle-GAN and Rev-GAN assume one-to-one mapping in image translation, ignoring diversities in transformation directions. MUNIT-GAN[40] (Fig. 1F) tackles this problem by first encoding the source data into one shared content space $C$, and one domain-specific style space $S$. The content code of the input is combined with different style codes in the target style space to generate target data with distinct styles.

## Evaluation Metrics

A set of metrics has been used for evaluating the quality of data generated by GAN-based models. Mean square error (MSE), peak signal-to-noise ratio (PSNR),[70] and structural similarity (SSIM)[70] were proposed for quantifying similarities or distances between paired data. Thus, they are typically used for comparing the generated data with the ground truth images. Specifically, MSE and PSNR measure the absolute pixel-wise distances between two images, while SSIM measures the structural similarity by considering dependencies among pixels. Two other metrics, Fréchet inception distance (FID)[156] and maximum mean discrepancy (MMD),[157] are utilized for computing distances between two data distributions when there is no paired ground truth. In the application of GAN-based models, they are often applied to measure similarities between distributions of generated and real data.

# GANs in Clinical Diagnosis

Accurate disease diagnosis is necessary for early intervention that may potentially delay disease progression. This is especially true in the case of neurodegenerative diseases, which are often highly heterogeneous, comorbid, and progress rapidly with severe impacts on the physical and cognitive function of patients. In the last decade, there has been pivotal progress in imaging techniques such as structural MRI, Fluorodeoxyglucose-PET (FDG-PET) and resting state functional MRI (rs-fMRI), enabling more precise and accurate measurement of disease-related structural and functional brain change *in vivo*. Simultaneously, advanced DL methods have been developed to analyze large, high-dimensional datasets and perform tasks such as disease classification and anomaly detection. GANs in particular have been leveraged to improve performance in both of these tasks.

This section is organized as follows. First, we discuss the use of GANs in disease classification frameworks with a primary focus on Alzheimer's disease. This is first discussed in the context of single modality imaging and then multimodal imaging. Next, we discuss the use of GANs in anomaly detection (see Table 2).

## Disease Classification

Classification involves clustering observations into distinct groups and assigning class labels to these groups based on associated input features. In the context of disease diagnosis, the classes could refer to disease stages or disease subtypes, for example. Previous work in disease classification relied on traditional ML techniques, such as SVM and logistic regression.[143,144] GANs can glean and analyze patterns in high-dimensional, multi-modal imaging datasets, detecting signs of neurodegenerative processes and underlying pathology at preclinical stages. They can also synthesize whole images across modalities, which can be used to assist classification downstream.

While GANs can be applied to many different disease datasets, most of the papers we discuss below focus on Alzheimer's disease, which is an irreversible neurodegenerative disease that debilitates cognitive abilities. It is the leading cause of dementia and currently impacts five million people in the United States.[148] GANs can derive powerful imaging markers for individualized diagnosis, classification into conversion groups, or even prediction of onset at preclinical and cognitively unimpaired stages.

## Disease classification with single-modality imaging

### Structural MRI

Structural imaging, including T1- and T2-weighted MRI, helps visualize brain anatomy such as shape, position, and size of tissues within the brain. Features that are commonly derived from structural MRI include regional brain volumes from T1-weighted scans and tissue hyperintensities from T2-weighted scans. Other brain characteristics such as tissue composition fractions and intracranial volume can also be extracted from these imaging modalities. Volumetric regional and whole brain atrophies derived from structural MRI are now identified as valid biomarkers of neurodegeneration, and have been used for clinical assessment and diagnosis.[149] Besides diagnostic utility, features extracted from structural imaging have been used as imaging endpoints to quantify outcomes in clinical trials of disease modifying therapies.[149]

Machine learning methods such as random forests have been proposed for automated Alzheimer's disease classification but these methods require careful feature extraction and selection.[150] Extracting features from structural MRI involves complex preprocessing steps, and the following feature selection phase requires advanced clinical knowledge. Supervised deep learning methods, such as convolutional neural networks (CNN), have also been proposed for Alzheimer's disease classification.[151,152] While CNNs can implicitly extract hierarchical features from images, they require large amounts of labeled training data, which are not always available.[151,152]

Therefore, Yu et al.[41] propose a 3D semi-supervised learning based GAN (THS-GAN) which utilizes both labeled and unlabeled T1-weighted MRI for classifying mild cognitive impairment and Alzheimer's disease. THS-GAN is modified based on C-GAN, and it has a generator, a discriminator, and a classifier. The schematic of the network architecture is shown in Fig. 2A. The generator uses 3D transposed convolutions to generate 3D T1-weighted MRI, while the discriminator and classifier use 3D DenseNet[42] to extract features from high dimensional MR volumes. Specifically, the generator which is conditioned on disease class produces a fake image-label pair, whereas the classifier takes an unlabeled image, predicts its corresponding disease category, and produces image-label pairs for unlabelled images. The discriminator's job is to identify whether an image-label pair comes from the real data distribution. When training three of them together, the generator tends to generate more realistic images for a given disease class, the classifier tries to improve its predictive accuracy, while the discriminator will maximize the probability of assigning fake labels to the image-label pairs generated from the classifier and generator. During evaluation, the model is able to learn and generate plausible images (as shown in Fig. 2B). The method achieves 95.5% accuracy in Alzheimer's disease v.s. healthy control classification, and 89.29% accuracy in mild cognitive impairment v.s. healthy control classification. We will discuss other GAN-based techniques that use structural imaging for disease classification in the multi-modal imaging section.

**Resting-state fMRI**

Rs-fMRI measures the time series of the blood-oxygenation-level-dependent (BOLD) fluctuations across brain anatomical regions. It relies on the underlying assumption that brain regions that co-activate, i.e. reliably demonstrate synchronous, low-frequency fluctuations in BOLD, are more likely to be involved in similar neural processes than regions that do not co-activate. Computing the Pearson correlations between the time series recorded in different brain regions provides estimates of functional connectivity, which is used to extract resting-state functional networks (rsFNs). These networks show patterns of synchronous activity across a set of distributed brain areas and provide potential biomarkers for a variety of illnesses.[44-46] Specifically, changes in the representation of rsFNs have been observed in groups suffering from brain disorders such as epilepsy, schizophrenia, attention deficit hyperactivity disorder and major depressive disorder, and diseases such as Alzheimer's and Parkinson's.[47-52] Prior works in fMRI-based disease classification rely on common machine learning techniques such as support vector machine (SVM) and nearest neighbors.[53] These techniques require careful feature engineering and feature selection to achieve optimal performance.

Recently, deep learning techniques such as convolutional neural networks have been used for disease classification based on automatic feature extraction.[54] However, due to insufficient data especially from rs-fMRI, these methods show poor generalizability. To overcome the limitations of both traditional ML and DL, researchers propose to use GANs to improve the classification performance. Zhao et al.[55] propose adapting a 2D GAN model for disease classification using rs-fMRI. They use a GAN architecture to classify individuals with mental disorders from healthy controls (HC) based on functional connectivity (FC). FC assesses temporal relationships of a subject's brain functional networks by computing the pairwise correlation between the spatially segregated networks.[56] Hence, FC reflects connectivity/synchronized activity of the brain networks and can be potentially used to identify fMRI-based biomarkers for disease classification. The generator takes a noise vector as input and learns to generate fake FC networks. The discriminator is used to classify mental disorders from HC, and discriminate between real and fake images. Finally, the model can be trained by optimizing an objective function that combines both adversarial loss and classification loss. The model performance is evaluated on two tasks, namely major depression disorder and schizophrenia classifications. The performance of the GAN model is validated against six classification techniques including k-nearest neighbors, adaboost, naive Bayes, Gaussian processes, SVM, and deep neural net. The GAN model outperforms all other methods in both tasks, i.e., major depressive disorder classification and schizophrenia classification, suggesting its utility as a potentially powerful tool to aid discriminative diagnosis.

Since the topology of brain connectomes is close to graphs, a natural extension would be boosting the classification model by using both graph structures and GANs. Mirakhorli et al.[57] use FC to identify abnormal changes in the brain due to Alzheimer's disease. The technique leverages FC to represent the human brain as a graph and then uses a graph neural network to learn structures which differentiate Alzheimer's disease subjects from healthy individuals. Here, a VAE which is implemented by graph convolutional operators serves as a generator and a discriminator is used to improve the recovery of the graphs. At inference, the encoder part of the VAE converts graph data into a low-dimensional space, and then abnormal signals (salience alteration of the brain connection) can be detected by comparing the differences of the graph properties (first-and second-order proximities) within these latent space features. The model achieves an average five-fold cross-validation accuracy of 85.2% for the three-way classification. The model also finds that abnormal connections of the frontal gyrus and precentral gyrus with other regions have a high percentage of Alzheimer's disease risk in the early stages and fall into the effective biomarkers category. Additionally, the olfactory cortex, supplementary motor area, and rolandic operculum have a high contribution to classify mild cognitive impairment patients. By recovering the missing connections with a generative approach and distinguishing the abnormal partial correlations from the healthy ones, the model provides biological-meaningful findings with high accuracy disease classification performance.

The studies mentioned above focus solely on functional connectivity. Even though performing classification using single-modality data from structural or functional MRI provides reasonable diagnostic accuracy, it can be boosted by using multi-modality data since additional modalities provide complementary information.

## Disease classification with multi-modal imaging

Multiple imaging modalities, such as MRI, PET, diffusion tensor imaging (DTI), and rs-fMRI, help in capturing diverse pathology patterns that may highlight different disease-relevant regions in the brain. This enhances the ability of disease classification models to distinguish diseases that are often comorbid, such as Alzheimer's and Parkinson's. However, the use of multi-modal imaging features is particularly challenging because of data sharing limitations, patient dropout, and relatively limited datasets with all modalities. Previous studies address this issue by simply discarding modality-incomplete samples.[58-61] This approach is prone to reducing classification accuracy and deteriorating the model's generalizability due to the limited sample size. Instead, GANs can better handle missing data in multi-modal datasets by generating the missing images and preserving sample size, thereby boosting downstream classification performance.

Previous work shows direct and indirect relationships between functional and structural pathways within the human brain.[62,63] These interesting studies have pivoted clinical research towards multi-modal integration to reliably infer brain connectivity. They also provided key insights into brain dysfunction in neurological disorders such as autism,[64] schizophrenia,[65] and attention deficit hyperactivity disorder.[66]

Pan et al.[67] propose to use multi-modal imaging to detect crucial discriminative neural circuits between Alzheimer's disease patients and healthy subjects. The model can effectively extract complementary topology information between rs-fMRI and DTI using a decoupling deep learning model (DecGAN). DecGAN consists of a generator, a discriminator, a decoupling module, and a classification module. The generator and discriminator modules capture the complex distribution of functional brain networks without explicitly modeling the probability density function. The decoupling module is trained to detect the sparse graphs which store relationships between region-of-interest connectivity, such that the classification module can accurately separate Alzheimer's disease and the healthy ones when taking these sparse graphs as inputs. The method shows accurate classification performance when discriminating HC v.s. early mild cognitive impairment (86.2% in accuracy), HC v.s. late mild cognitive impairment (85.7% in accuracy) and HC v.s. Alzheimer's disease (85.2% in accuracy). The model also finds that limbic lobe and occipital lobe are highly correlated to Alzheimer's disease pathology.[68,69] One major limitation of this work is that the authors assume the coupling between two regions is static, which conflicts with several recent studies that show functional connectivity is dynamic.[153-155] Moreover, the sample size of the study is small (236 subjects), which hinders the generalizability and reproducibility of region-of-interests detected by the model. Future studies could incorporate dynamic connectivity and detect dynamic changes predictive of Alzheimer's disease or other neurodegenerative diseases and disorders.

Lin et al.[33] use a 3D Rev-GAN[39] for missing data imputation and then evaluate the effect of the inclusion of GANs-generated images in Alzheimer's disease v.s. cognitively unimpaired (CU) as well as stable mild cognitive impairment v.s. progressive mild cognitive impairment classification. The method is evaluated on CN subjects, subjects with stable mild cognitive impairment and progressive mild cognitive impairment and subjects with Alzheimer's disease. Hippocampus images are used in addition to the full brain images in the experiments. Rev-GAN constructs synthetic PET images with high image quality that slightly deviates from real PET scans. Compared to other image synthesis methods that perform more processing steps to achieve higher alignment between the different modalities,[71-74] this approach yields comparable PSNR and higher SSIM in PET images synthesis. In terms of MR hippocampus images synthesis, the Rev-GAN achieves the highest SSIM and PSNR. The model performance drops for the full image reconstruction due to the difficulties in mapping the structure information such as the skull of the MR image from the functional image. Overall, the use of only one generator to perform bidirectional image synthesis in combination with the stability of reversible architecture enables the training of deeper networks with low memory cost. Therefore, the non-linear fitting ability of the model is enhanced, resulting in the construction of high quality images (see Fig. 2C).

After imputing the missing data with GANs, Alzheimer's disease diagnosis and mild cognitive impairment to Alzheimer's disease conversion prediction are implemented using a multi-modal 3D CNN. The model trained using real hippocampus images for one modality and fully synthetic data from the other modality yields similar, sometimes superior, performance compared to the model using real data for both modalities, and always higher performance than the model using missing data. In terms of full images, although the quality of the generated MR full images is not as good as that of the generated hippocampus images, the classification accuracy using synthetic MR full images exceeds 90% for the Alzheimer's disease diagnosis and 73% for the mild cognitive impairment to Alzheimer's disease conversion prediction. Overall, the prominent improvement of the classification results with the use of GAN-synthetic data reveals the ability of the image synthesis model to construct images of high quality which also contain useful information about the disease, thus significantly contributing to Alzheimer's diagnosis and mild cognitive impairment conversion prediction.

In this study, the authors use missing data synthesis to improve the Alzheimer's disease diagnosis and the prediction of mild cognitive impairment to Alzheimer's disease conversion. However, less attention has been devoted to the FDG metabolic changes and the biological significance of the imputed data compared to real data. Additionally, a prerequisite for successful MRI-to-PET mapping is that the disease affects the tissue structure and metabolic function at the same time. Further exploration of the MRI-PET relationship in diseases such as cancer where structural and functional changes do not occur simultaneously is needed.

To date, relatively less attention has been devoted to the generation of amyloid PET images. Amyloid PET measures the amount of amyloid beta protein aggregation in the brain,[147] which is one of the key hallmarks of Alzheimer's disease. However, the availability of PET scans is extremely limited (compared to the MR scans) due to the radioactive exposure and high cost. Yan et al.[34] use a 3D conditional GAN to construct $^{18}$F-florbetapir PET images from MR images and then compare the performance of their method with traditional data augmentation techniques, such as image rotation and flipping, in a mild cognitive impairment classification task. The conditional GAN generator is a U-Net[75] based CNN with skip connections and the discriminator is PatchGAN discriminator.[38] The generator is not only trained to fool the discriminator but also to construct images as close to reality as possible. Paired PET-MRI images are used to train the C-GAN. Then, the trained model is applied to generate PET images from MR images and the real PET images are used for the evaluation of C-GAN using SSIM metric. The SSIM reaches 0.95, thus indicating the ability of the model to generate images with high similarity with the real images. To classify stable mild cognitive impairment v.s. progressive mild cognitive impairment, a residual network (ResNet)[76] is built. The ResNet is trained using PET images from three scenarios: real PET images only, combined real PET and PET images generated using traditional augmentation techniques, and combined real PET and C-GAN-generated PET images. The classification performance increases with the aid of synthetic data. Between the two approaches, the inclusion of C-GAN-generated PET images in training results in higher performance compared to the model that used images generated using traditional augmentation techniques and this reveals the superiority of GANs in image synthesis over traditional image augmentation techniques. Medical images are different from the natural ones with a certain centering, alignment and asymmetry geometry, as well as characteristics such as contrast and brightness. Thus, computer vision augmentation techniques, such as adjusting brightness or contrast, adding noise, might alter the semantic content of the image; for example, the distinction between gray and white matter tissues can be impeded when changing the contrast of the image.

Hu et al.[141] extend the MR-to-PET synthesis framework by developing a 3D end-to-end network, called bidirectional mapping GAN (BMGAN). This model adopts 3D Dense U-Net, a variant of U-Net[94] that leverages the dense connections of DenseNet[42], as the generator to synthesize brain PET images from MR images. The densely connected paths between layers in DenseNet tackle the vanishing gradient problem, foster feature propagation and information flow, and reduce the number of network parameters. One advantage of BMGAN is that it sets up an invertible connection between the brain PET images and the latent vectors. The model does not only learn a forward mapping from the latent vector to the PET images as traditional GANs, but also learns a backward mapping that returns the PET images back to the latent space by training an encoder simultaneously. This mechanism enables the synthesis of perceptually realistic PET images while retaining the distinct features of brain structures across individuals. Beside the high-quality generated images, the effectiveness of these images in disease diagnosis has been demonstrated by performing Alzheimer's disease v.s. normal classification. The classification performance

(AUC) using the BMGAN synthesized PET images is better than those generated by state-of-the-art medical imaging cross-modality synthesis models such as CGAN and PGAN.[142]

**Joint image synthesis and classification paradigm**

Many imputation methods for multi-modality neuroimaging datasets usually treat image synthesis and disease diagnosis as two separate tasks.[33,34,67] This ignores the fact that different modalities may identify different relevant regions in the brain relevant to the disease being studied. Performing image synthesis and classification in a joint framework enables deep learning networks to leverage correlations across input modalities.

Gao et al.[24] propose a 3D task-induced pyramid and attention GAN (TPA-GAN) to generate missing PET data given the paired MRI. The pyramid convolution layers can capture multilevel features of MRI while the attention module eliminates redundant information and accelerates convergence of the network. The task-induced discriminator helps generate images that retain information specific for disease classification. Then, a pathwise-dense CNN (PT-DCN) gradually learns and combines the multimodal features from both real and imputed images towards the final disease classification. The pathwise transfer blocks consist of a concatenation layer, convolution layer, batch normalization and ReLU activation layer, and a larger convolution layer. These blocks are used to communicate information across the two paths of PET and MRI, making full use of complementary information in these two modalities. Under SSIM, PNSR and MDD metrics, the TPA-GAN outperforms several baseline methods, including a CycleGAN variant developed by Pan et al.,[71] for generation of PET images using MRI. In the experiments, the authors use ADNI-1 for training and ADNI-2 for testing, which could be an issue if the imaging data is not harmonized appropriately across scanner-changes, acquisition protocols, and subject demographics. In the future, cross-study transfer learning or domain adaptation techniques can be investigated to alleviate the problem. The following work leverages this idea to improve the power of a sample-size limited clinical study.

Pan et al.[77] extend on the joint synthesis-classification method developed by Gao et al.[24] by maximizing image similarity within modalities. They propose a disease-image-specific deep learning (DSDL) framework for joint neuroimage synthesis and disease diagnosis using incomplete multi-modality neuroimages. First, disease characteristics specific to a given image modality are implicitly modeled and output by a disease-image-specific network (DSNet), which takes whole-brain images as input. A feature-consistency GAN (FGAN) then imputes the missing images. The FGAN encourages feature maps between pairs of synthetic and real images to be consistent while preserving the disease-image-specific information, using the outputs generated by DSNet. Therefore, the FGAN is correlated with DSNet and synthesizes the missing modalities in a diagnosis-oriented manner, resulting in better performance. Specifically, the DSNet achieves an diagnostic performance of 94.39% with only MRI and 94.92% with MRI and PET when using AUROC as the metric.

The joint neuroimage synthesis and representation learning (JSRL) framework proposed by Liu et al.[78] offers a few advantages compared to the previous works. The model integrates image synthesis and representation learning into a unified framework where the synthesized multimodal representations are used as inputs for representation learning. The framework leverages transfer learning for prediction of conversion in subjects with subjective cognitive decline, which is the self-reported experience of worsening confusion or memory loss. JSRL consists of two major components: a GAN for synthesizing missing neuroimaging data, and a classification network for learning neuroimage representations and predicting the progression of subjective cognitive decline. These two subnetworks share the same feature encoding module, encouraging the generated data to be prediction-oriented. The underlying association among multimodal images can be effectively modeled for accurate prediction with an AUROC of 71.3%. In summary, this method focuses on improving the classification of subjective cognitive decline subjects using incomplete multimodal neuroimaging data. Since subjective cognitive decline is one of the earliest noticeable symptoms of Alzheimer's disease and related dementias, the classification is clinically useful to begin targeted interventions earlier in these subjects. This work is among the first multimodal neuroimaging-based studies for subjective cognitive decline conversion prediction, which avoids the need to individually extract MRI and real or synthetic PET features as in previous works. JSRL leverages transfer learning by harnessing a large scale ADNI database to model a smaller scale database on subjective cognitive decline, which significantly increases the power of this study.

While many studies apply GANs for image synthesis and classification in neurodegenerative diseases, GANs have broader application in neurology, including detection of brain tumors and imaging anomalies. These applications are discussed in the next section.

## Tumor and Anomaly Detection

### Supervised tumor detection

Brain tumors, abnormal proliferations of cells in the brain, comprise a large portion of deaths related to cancer worldwide.[145] Tumor detection and classification is an active research area in the medical imaging community; however, available imaging data for this research purpose remains relatively limited.[146] To tackle this problem, many of the following recent works leverage the generative abilities of GANs for dataset enrichment and augmentation.

Han et al.[23] demonstrate the use of GANs in improving the performance of a brain tumor detection network. They propose a two-step method for enriching the training dataset via data augmentation by generating additional samples of normal and pathologic images. They use an initial 2D noise-to-image GAN to produce the anatomical content and rough attributes of a scan and sequentially an unpaired image-to-image translation network to refine these images.

Park et al.[80] utilize StyleGAN to create synthetic images while preserving the morphologic variations to improve the diagnostic accuracy of isocitrate dehydrogenase (IDH) mutant gliomas. The 2D GAN model was trained on normal brains and IDH-mutant high-grade astrocytomas to generate the corresponding contrast-enhanced (CE) T1-weighted and fluid-attenuated inversion recovery (FLAIR) images. The authors further develop a diagnostic model from the morphologic characteristics of both realistic and synthetic data to validate that the synthetic data generated by GAN can improve molecular prediction for IDH status of glioblastomas.

While these approaches show initial promise, GAN-based dataset enrichment has not been thoroughly studied for brain tumors. For example, GANs might not be able to capture the sample distribution of highly heterogeneous tumor data when only limited data is available. Given this challenge, GAN-based unsupervised anomaly detection offers distinct advantages for tumor detection.

## Anomaly detection

Anomaly detection, the identification of scans that deviate from the normative distribution, offers a path toward identifying pathology even when the anomalous group is not explicitly defined. Anomaly detection has most often been used for the detection and segmentation of tumors and lesions.[84,86] Prior approaches for anomaly detection are not suited for use at the image level and often use regional summary measures from segmented brain regions to identify abnormalities from scans. For example, one-class SVM is used to define the normative group allowing outliers to be identified.[81,82] These methods have shown promise in specific cases but are sensitive to the selection of summary measures used to represent the scan. In general, anomaly detection models are trained using both healthy and anomalous brain scans or healthy scans only. Focusing more on the latter case, GAN-based anomaly detection in neuroimaging stems from the ability of GANs to model the normative distribution of brains accurately. When substantial deviations from the expected distribution occur, the model can infer the presence of abnormalities, which has been leveraged in neuroimaging for lesion and tumor detection. In 2017, AnoGAN[83] gained popularity as an anomaly detection method using only normative samples to define its detection criteria. AnoGAN uses a generator to learn the mapping from a low-dimensional latent space to normal 2D images, defining normal/healthy regions in the latent space. When a new image is encountered, the latent representation whose reconstruction matches the new image most closely would be selected via backpropagation-guided sampling. Since the generator is only trained on normal samples, the learned latent space cannot adequately represent the variation of anomalous scans, and thereby the reconstructed images from anomalous images often differed in the anomalous regions (see Fig. 3). If deviation between the reconstructed and original image is observed, the image would then be marked as anomalous.

Similarly, Nguyen et al.[84] develop an unsupervised brain tumor segmentation/detection method leveraging GAN-based image in-painting technique – the reconstruction of partially obscured areas of an image from the surrounding context. If the network is trained on healthy images, anomalous regions of an image will be in-painted. The method performed well in images with smaller, local anomalies as the surrounding context contained enough information for the in-painting of a healthy region. Training in this scenario is performed by randomly masking a part of an image and asking the generator to recreate the missing portion based on the unmasked regions. During inference, the target image is masked in many different positions and then reconstructed by the network. If there exists deviation between the reconstructed and original images in several subsets of the masked images, this subject is likely to be marked as an anomaly. By repeatedly masking various parts of the scan, the authors are able to generate a tumor segmentation mask. The authors report an improvement of tumor segmentation performance over AnoGAN with a Dice score from 38% to 77%. Bengs et al.[86] implement a similar idea of unsupervised anomaly detection to detect brain tumors using VAE-based image in-painting. The authors demonstrate an improvement over prior 2D methods in brain tumor segmentation with a Dice score from 25% to 31%. Although image in-painting methods have shown promise for abnormality detection, they have two major limitations. Firstly, they do not perform well when the abnormality is large or has a global effect on the scan; secondly, the algorithms may produce false positives in regions where normal anatomical variation is high, because there might be multiple acceptable ways of in-painting a region only based on its surrounding appearance.

GAN-based methods have offered a unique way to identify deviations from the healthy distribution. In particular, GAN-based unsupervised anomaly detection shows promising performance when pathology data is limited and difficult to acquire. There are a variety of potential clinical and research-based applications for such methods. It is foreseeable that these methods will be useful in triaging scans, with critical or time sensitive pathology, for a radiologist to read. While most published works have focused on identifying gross abnormalities, such as stroke and tumor lesions, it remains to be seen how well similar approaches perform in identifying subtler pathology. Within the domain of anomaly and tumor detection, GANs have also shown promise in enriching training data in cases with limited or missing data.

## GANs in Brain Aging and Modeling Disease Progression

To understand how a patient's brain changes due to pathology, it is important to first understand how the brain evolves in the absence of pathology. This motivates the need for methods that specifically tackle the problems with modeling healthy brain aging, and how GANs can be utilized to simulate subject specific brain aging.[88] Additionally, modeling abnormal disease specific brain changes is also of clinical significance. Disease progression modeling can help screen for people at risk of developing neurological conditions and also help plan preventative measures or treatment options. This section will discuss the challenges in modeling healthy and abnormal brain changes and motivate the need and utility of GANs in disease prognosis. We will also review various existing GAN models that are designed to model healthy and abnormal brain changes (see Table 3).

## Brain Aging

The human brain undergoes morphological and functional changes with age. Deviations from these normative brain changes might be indicative of an underlying pathology. Neuroimaging techniques such as structural and functional MRI have been successfully applied[89] to measure and assess these brain changes. Modeling brain aging trajectories and simulating future brain states can be valuable in a number of applications including early detection of neurological conditions and imputation for missing data in longitudinal studies. In prior work, researchers developed common atlas models[90-92] as spatio-temporal references of brain development and aging. One of the main challenges with this approach is that individuals might exhibit unique brain aging trajectories based on their lifestyle and health status. However, common atlas models might not preserve this inter-subject variability resulting in inaccurate modeling. In recent years, GANs have been proposed to combat this issue and generate subject-specific brain aging image synthesis.

Xia et al.[88] designed a conditional GAN that synthesizes subject-specific brain images given a target age and health condition. Their unique model learns to synthesize older brain MR scans from a subject's current brain scan without relying on any longitudinal scans to guide the synthesis. As depicted in Fig. 4A, their model consists of a generator that is conditioned on the target health state and the difference between current age and target age of the subject. The generator takes a 2D T1-weighted MRI and synthesizes brain images that correspond to target age and health state (control/mild cognitive impairment/Alzheimer's disease) while preserving subject identity. On the other hand, the discriminator ensures that the generated images correspond to the target age and health state by learning the joint distribution of the brain image, target age, and target health state. To preserve individual brain characteristics of the subjects during modeling, the authors train the GAN model with a combination loss function that has three elements: an adversarial loss ($L_{GAN}$ is a Wasserstein loss with gradient penalty) that encourages the model to generate realistic brain images, an identity-preservation loss ($L_{ID}$) that encourages network to preserve the subject's unique characteristics during image generation, and finally a reconstruction loss ($L_{rec}$) that encourages the network to reconstruct input when the generator is conditioned on the same age and health state as the input. Fig. 4B shows the results of the conditional GAN in synthesizing images of the brain at multiple target ages. Although the predicted apparent age of synthesized images in Fig. 4B is very close to the target age, one limitation of the method is that the subject identity might not be preserved during image synthesis. The authors only incorporate age and health state in the modeling process, but other factors such as gender and genotype can help model finer subject details that might help preserve subject identity. Additionally, the model was trained to synthesize older brain scans from younger brain scans but not vice-versa. Modeling the opposite will not only strengthen the usability of the model for imputing missing timepoints but also provide a more robust model that preserves subject identity. Another major limitation is that the model uses a 2D design, to improve visual quality of the generated brain images, 3D architectures can be adapted to model the brain as a whole volume. The subsequent paper tackles some of these limitations.

Unlike the 2D model presented by Xia et al.,[88] Peng et al.[93] introduce 3D models that longitudinally predict brain volumes in infants during their first year of life. The first model they introduce is a single-input-single-output model called perceptual adversarial network (PGAN). As depicted in Fig. 4C, PGAN has a 3D U-Net[94] as the generator which aids in volumetric processing. The generator takes T1 or T2 weighted brain images from an initial timepoint and learns to generate corresponding longitudinal brain images. The discriminator learns to discriminate the fake images from the real images using adversarial loss $L_{GAN}$. Additionally, a voxel-wise reconstruction loss $L_{VR}$ encourages the voxel intensities of the generated images to be close to the corresponding voxel intensities of the ground truth images. Since the voxel-wise reconstruction loss might over-smooth the generated images an additional loss function called perceptual loss $L_P$ is introduced. Perpetual loss helps preserve the sharpness of the generated images. Since MRI sequences capture complementary features of the brain, the authors propose a second model called multi-contrast perceptual adversarial network (MPGAN). This model extends the PGAN architecture to incorporate multiple modality inputs and outputs, thereby learning complimentary features from both T1- and T2-weighted brain images. As depicted in Fig. 4D, MPGAN has two generators based on a 3D U-Net architecture, but unlike PGAN, the 3D U-Net has a shared encoder that takes T1- and T2-weighted images at a given time point as input, and two independent decoders that synthesize the longitudinal T1- and T2-weighted brain images respectively. Two discriminator networks learn to discriminate between real and fake T1- and T2-weighted images. The models are evaluated on an infant brain imaging dataset with T1- and T2-weighted volumes available at 6 months and 12 months of life. The model performance is measured on two tasks: predicting six month images from twelve month images as well as predicting twelve month images from six month images. A major limitation of this work is that it's constrained to modeling two timepoints since paired images are used to train and model brain aging at 6 and 12 months of life. Hence the model cannot synthesize brain images over the whole lifespan, to do so would require scanning subjects across their whole lifespan. This leads us back to a recurring problem in modeling the brain aging process: models might require longitudinal data for training, which may be infeasible to acquire.

Brain aging is a complex process and each individual presents a unique brain aging trajectory that is influenced by their age, genetic code, demographics, and any underlying neuropathy. GANs allow for subject specific synthesis of the aging brain, but there is no guarantee that the subject identity is preserved during synthesis. Hence, future research needs to focus on integrating multiomic, imaging, and clinical data for brain aging synthesis while ensuring the preservation of subject identity.

## Alzheimer's Disease Progression

The human brain deviates from normative brain aging when underlying disease processes affect its structure and function. Disease progression models trained on longitudinal imaging data can characterize the future course of the disease progression, making them valuable for clinical trial management, treatment planning and prognosis. Traditional ML algorithms have been widely applied for modeling Alzheimer's disease progression, with a focus on extrapolating biomarker metrics and cognitive scores. For example, Zhou et al.[95] propose a least absolute shrinkage and selection operator (LASSO) formulation to predict Alzheimer's disease patients' cognitive scores at different time points. Recently, disease progression modeling is not only approached as a regression task, but also a generative task where models generate realistic high-dimensional image data. Researchers leverage the ability of GANs to synthesize realistic images and other data in general, in order to simulate future states of Alzheimer's disease.[22,30,36] Generating realistic high dimensional data in the medical field is far from being considered a trivial task, due to the complexity and irregular availability of longitudinal and annotated data. GANs have been predominantly used in disease progression modeling because of their superior ability in learning sharp distributions from training data and producing high resolution images, compared to other generative techniques.[22,30,36,88,93]

For example, Ravi et al.[36] present a 4D-degenerative adversarial neuroimage net (4D-DANI-Net), with the goal of generating high resolution, longitudinal 3D MRIs that mimic the personalized neurodegeneration using spatiotemporal and biologically-informed constraints. The 4D-DANI-Net is composed of three main blocks:a preprocessing block, a progression block, and a 3D super resolution block. The preprocessing blockremoves irrelevant variations in the data. The progression modeling block is implemented using the degenerative adversarial neuroimage (DANI) net[96] (see Fig. 5A). A conditional autoencoder (CAD), a set of adversarial networks and a set of biological constraints are the main elements of DANI net. The CAD is responsible for producing the longitudinal 2D MRI with the use of the biological constraints in the optimization and a discriminator that is going to compare the fake longitudinal data produced by the CAD with the real ones. Multiple DANI-Nets are trained, one for every 2D MR slice, and all these DANI nets compose the progression model of the 4D-DANI-Net. In order to unify the low resolution images produced by the DANI nets, the authors employ the 3D super-resolution block that produces the 3D high resolution MR image. The final block is the super resolution one that transforms the low-resolution images, produced by the DANI nets, into the 3D high resolution MR image. In Fig. 5B, there are qualitative results that showcase the ability of 4D-DANI-Net to produce MR scans over time that correspond in different ages of the same subject.

In the same spectrum of simulating Alzheimer's disease progression but in 2D space, Bowles et al.[22] built a progression model for Alzheimer's disease that leverages the imaging arithmetic and isolates the features in the latent space that correspond to Alzheimer's pathology. The core model is W-GAN along with a re-weighting scheme. The re-weighting scheme increases the weighting of those real images that are misclassified. This forces the discriminator to better represent the most extreme parts of the images, which in turn forces the generator to produce images from this region. Using imaging arithmetic and the latent encodings that correspond to Alzheimer's disease features one can simulate scans with Alzheimer's with different grades of severity. For example, by adding the latent encoding of Alzheimer's, with a specific scaling, in the MR scan of a cognitive normal subject one can see enlarged ventricles and cortical atrophy in the output MRI. However, this paper makes several potentially problematic assumptions. They have assumed that the progression is a linear process over time. Furthermore, they hypothesized that morphological changes across all subjects with Alzheimer's disease are symmetric. Additionally, this methodology is developed using a small window size, 64 by 64 which therefore makes it unrealistic for a use case scenario.

Yang et al.[30] present an alternative GAN-based approach for Alzheimer's disease progression analysis built on tabular volumetric data. Firstly, they worked on disentangling the structural heterogeneity of the diseased brain and then with meta-analysis connected the cross-sectional patterns to longitudinal data. They propose semi-supervised clustering GAN (SMILE-GAN), a method that manages to disentangle pathologic neuroanatomical heterogeneity and define subtypes of neurodegeneration. In general, SMILE-GAN learns mappings from cognitive unimpaired individuals to dementia patients in a generative approach. Through this approach, SMILE-GAN captures disease effects that contribute to brain anatomy changes and avoids learning non-disease related variations such as covariate effects. Technically, the model learns one-to-many mappings from the CN group, $X$, to the patient (PT) group $Y$. The goal is to learn a mapping function $f: X \times Z \rightarrow Y$ which generates fake PT data from real CN data. The subtype variable $z$ is used as an additional input to the mapping function $f$, along with the the CN scan $x$. Along with the mapping function, the model also trains the discriminator $D$ to distinguish real PT data $y$ from synthesized PT $y'$. The optimal number of clusters is determined using cross-validation and evaluating for reproducibility of the results. These clusters define a four-dimensional coordinate system that captures major neuroanatomical patterns which are visualized in Fig. 6B.

This work goes on to connect the identified clusters with longitudinal progression pathways. Yang et al.[30] provide an alternative and robust way to model progression, by identifying patterns of atrophy. No assumptions on imaging data distribution and its independence from confounding factors and variability make it a powerful model that could potentially be used in subtyping of other heterogeneous diseases. Furthermore, with more representative data such a method can potentially identify more intricate subtypes that now are covert due to its limited instances in the current datasets.

Modeling the progression of Alzheimer's is a challenging task due to the complexity, availability, and the multiple modalities, ranging from imaging ones such as MRI and PET to genomic and clinical information. Understanding the underlying biological processes and trying to comprehend potential factors that connect these modalities is the way to interpret and build knowledge for heterogeneous diseases such as Alzheimer's. A future challenge and research opportunity is to develop GAN models incorporating high resolution genomic information as single nucleotide polymorphisms (SNPs). Exploring a pathway between genomic data and imaging signatures will shed light on Alzheimer's endophenotypes and such knowledge is valuable for potential future treatments.

## Progression of Brain Lesions

MRI-visible brain lesions, such as white matter hyperintensities and multiple sclerosis lesions, reflect white matter or gray matter damage caused by chronic ischaemia associated with cerebral small vessel disease or inflammation that results from the malfunction of the immune system.[97,98] Since white matter hyperintensities play a key role in aging, stroke, and dementia, it is important to quantify white matter hyperintensities using measures such as volume, shape and location. These measures are associated with the presence and severity of clinical symptoms that support diagnosis, prognosis, and treatment monitoring.[99] We can observe the hyperintense regions clearly in T2-weighted and FLAIR brain MRI. The evolution of white matter hyperintensities over a period of time can be characterized as volume decrease (regress), volume stability, or volume increase (progress). It is challenging to predict the evolution of white matter hyperintensities because its influence factors such as hypertension and aging are poorly understood.[100] White matter hyperintensities evolution prediction is under-explored in the literature, though other lesion progression, e.g. ischemic stroke lesion, has been modeled using a non-linear registration method called longitudinal metamorphosis.[101]

Rachmadi et al.[102] predict the evolution of white matter hyperintensities by generating a "disease evolution map" using 2D GANs. In the study, the baseline and follow-up scans are represented with irregularity maps which describe the abnormal level in voxel resolution. Further, the disease evolution maps are obtained by subtracting the baseline irregularity maps from the follow-up irregularity maps. To model the evolution process of white matter hyperintensities, a generator, which is implemented with an autoencoder namely U-Net,[75] is introduced to project an irregularity map at baseline to its corresponding disease evolution map in the follow-up year. A discriminator is used to classify if the disease evolution map came from the generator or the real scans. To enforce anatomically realistic modifications to the follow-year irregularity maps, another discriminator is introduced to identify the irregularity maps at follow-up year from those follow-year generated maps which is obtained by summing up the generator input and output. The model schematic is shown in Fig. 7A. The GAN model can accurately predict a subject's white matter hyperintensities volume in the follow-up year with 1.19 ml error in average, and its classification accuracy on whether each white matter hyperintensities subject will regress or progress is 72% without accessing any ground-truth labels. The qualitative assessment of the disease evolution maps are shown in Fig. 7B with red color indicating progress and blue color indicating regress. Since this is a very early attempt in tackling white matter hyperintensities dynamic prediction, there are several limitations in this work. Firstly, the model requires manual white matter hyperintensities labels from both baseline and follow-up sessions, which is not practical and tedious for human experts. Secondly, the method depends on the quality of the extracted irregularity map which tends to overestimate white matter hyperintensities in the optical radiation and underestimate it in the frontal lobe.

Multiple sclerosis is one of the most prevalent autoimmune disorders which affects the central nervous system.[103] With a relatively young age onset, multiple sclerosis has multiple symptoms such as vision loss, dizziness, and cognitive decline.[103] To better understand the physiopathology of multiple sclerosis, PET with [11C] Pittsburgh Compound-B radiotracer has been proposed to visualize and measure the myelin which insulates and protects the axon in the central nervous system, loss and repair in multiple sclerosis lesions.[104] Specifically, the loss of myelin surrounding the axon leads to the axon degeneration called demyelination. On the contrary, new myelin sheath generation can repair the damaged axon, namely remyelination. However, PET is an invasive and expensive imaging technique which is only available in limited hospitals around the world. Recently, a lot of interest has been drawn in predicting the PET-derived myelin evolution in multiple sclerosis from non-invasive and low-cost MRI. Although many studies[105,106] utilize traditional machine learning tools, such as structured random forest, for image modality prediction and artifact reduction, they seldom focus on the underlying pathology dynamics. Wei et al.[35] propose to predict the dynamic of myelin content changes, represented by the distribution volume ratio parametric map, using a two-stage conditional 3D GAN with attention regularization for multiple sclerosis lesions. At the first stage, a conditional GAN is utilized to map noise variable sampled from standard Gaussian distribution to PET image domain, where both the generator and the discriminator are conditioned on four multi-sequential MRIs from the same patient, including magnetisation transfer ratio map (MTR) and three DTI measures: fractional anisotropy (FA), radial diffusivity (RD), and axial diffusivity (AD). Next, the second stage generator takes in both the multi-sequential MR images and the output from the first stage generator and produces a refined PET image. The other discriminator tries to distinguish between the generator output and the ground-truth PET image conditioned on the four multi-sequential MR images. To model the spatially sparse lesion relationships, self-attention layers have been incorporated in the network architecture. The detailed model structure is shown in Fig. 7C. To validate the network, PET images of 18 multiple sclerosis patients are generated from both baseline and follow-up multi-sequence MRIs. Compared with the true longitudinal [$^{11}$C] PIB PETs, the Dice coefficient for the masks of demyelination and remyelination voxels in the GAN generated predictions are 71% and 69% on average separately. Visual assessments of the lesional myelin content changes are demonstrated in Fig. 7D. However, there are several shortcomings in this work. Firstly, the lesion segmentation masks used in the model are manually pre-defined, which is not practical for clinical usage. Secondly, the method utilizes multi-sequence MRIs as input for better PET synthesis quality. Incomplete inputs will cause performance degradation. Finally, the model has been only validated on a small, single-center dataset.

In this section, we introduced GAN-based tools which are developed for brain lesion progression estimation. Although these studies show promising potential in accurately predicting the dynamics of white matter hyperintensities and multiple sclerosis lesions with generative models, they suffer from problems in real-world deployment. For example, manually annotated lesion masks should be replaced by automated segmentation algorithms to reduce human labor cost. Additionally, longitudinal analysis benefits from the rich information from multi-modal data. In practice, however, the scarcity of complete multi-sequence MRIs for each patient induces another challenge. Finally, reproducibility and replicability are huge issues for machine learning research in small-scale datasets. We will provide potential solutions to these shortcomings in the discussion section.

## Brain Tumor Growth

Glioblastoma multiforme tumors, the most common and aggressive type of primary brain tumors, have high intra-tumor heterogeneity, leading to treatment failure and reduced survival.[138] Medical imaging techniques such as MRI, DTI, and PET are commonly used in clinical practice for the detection, diagnosis and examination of gliomas. Leveraging multiple time points of these imaging modalities can provide rich information for predicting tumor evolution. Importantly, the study of brain tumor growth is critical for the clinical diagnosis of the disease, particularly for treatment planning, tumor aggressiveness quantification and progression prediction.[139,140]

Previous work on brain tumor growth modeling mainly rely on complex mathematical formulations such as a system of partial differential equations to capture the effects of the invasion and diffusion of tumor cells.[107] However, these models are often insufficient to represent various growth processes due to their limited number of parameters. Additionally, the estimations of model parameters are also difficult for each individual without leveraging broader knowledge of tumor patterns in the cohort. A Bayesian approach has also been proposed by Angeli et al.[108] for the prediction of patient-specific tumor cell density from MRI and PET modalities. But this work was designed for personalized radiotherapy treatment planning by only using the cross-sectional dataset instead of longitudinal dataset, which does not enable the investigation of tumor growth pattern as time progresses. Deep learning-based methods have also been developed for tumor growth studies. For example, Zhang et al.[109] propose a spatial-temporal convolutional long short-term memory method to capture the temporal dynamics of pancreatic tumors from multiple time points of CT images, while the method has not been applied to study cerebral tumors. To address the above limitations, GANs have been leveraged for its ability to generate realistic data and have been lately applied in the prediction of brain tumor growth.

Elazab et al.[110] propose a novel method using stacked 3D GANs (GP-GAN) for the growth prediction of glioma. The framework consists of $n-1$ stacked GANs where $n$ denotes the number of time points. The generators are initialized by tumor boundaries and tissue feature maps (FMs), while FMs have semantic labels for white matter, gray matter and CSF. As shown in Fig. 8A, within each GAN block, T1 MR images and FMs are fed into the generator $G_i$ for every time point $t_i$. The FMs can help guide the generator to better estimate the tumor boundary at the next time point. Meanwhile, the discriminator $D_i$ is trained to distinguish between the generated tumor boundary and the ground truth. The entire GP-GAN can thus be progressively trained in an end-to-end fashion for jointly training all the network parameters. The training will enable generator $G_{i+1}$ to construct a good estimation of the tumor at the next time point from the current input. The objective function for training the network includes two loss terms as Dice and $L_1$ norm losses, which are beneficial for the generated images to be more similar to the ground truth at $t_{i+1}$ and for the model to be less affected by artifacts. GP-GAN further adapts a modified 3D U-Net architecture for the generators, which has the advantage of skip connections to integrate hierarchical features to better generate the images. Fig. 8B shows growth prediction results for subjects with low-grade glioma and high-grade glioma at different time points via GP-GAN. GP-GAN outperforms the other state-of-the-art reaction diffusion-based and deep learning-based tumor growth modeling methods, as measured by Dice coefficients and Jaccard index of 78.97% and 88.26%. For clinical applications, this method can potentially facilitate the computer-aided prognosis of brain tumors, while more detailed investigations are required, such as in vivo experiments for validation. However, there exist noticeable limitations to this method. Firstly, the current tumor growth study may not develop a more refined model for various segmented parts of the tumor, such as edema, tumor core and necrosis. This detailed tumor modeling by considering different tumor regions could generate more accurate and clinically relevant tumor growth prediction since different tumor parts could exhibit distinct growth patterns. Additionally, the refined model could potentially bring in a larger number of parameters, as required by modeling different tumor regions. This will increase the difficulty of training and the likelihood of overfitting, especially on medical imaging datasets with limited sample size.

Due to privacy concerns and the obstacles in establishing a large data consortium across multiple institutions, data shortage is another major challenge in studying brain tumor growth. Since GANs are commonly used to generate realistic data, Kamli et al.[111] propose a synthetic medical image generator (SMIG) based on 3D GAN to generate anonymised MRI for data augmentation. The objective of the proposed model is to tackle the issues of data privacy, imbalanced data and insufficient training samples, which could result in poor performance in terms of classification accuracy and sensitivity. In particular, SMIG creates different types of synthetic MR images, e.g., the generation of the abnormal brain based on the healthy brain and tumor volume, as well as changing tumor to a new location from the input of original image and tumor volume, as shown in Fig. 8C. Thus, SMIG is superior to traditional data augmentation techniques by providing different images with rich tumor information instead of applying geometric transformation to the original image. Fig. 8D shows examples of the application of the SMIG model on single patient images from the BraTS dataset. Additionally, the authors develop a tumor growth predictor (TGP) model based on an end-to-end CNN for tumor growth prediction. The prediction model is motivated by the lack of deep learning-based models for investigating tumor volume growth prediction as well as the availability of synthetic data from SMIG. The TGP network takes the first patient scan as the input and predicts the scan after 90 days (i.e., tumor volume). The network architecture consists of an encoder-decoder framework. The encoder extracts features into a low-dimensional latent representation, and the decoder produces the final output as the same size of input. The results demonstrate that a high accuracy for the tumor prediction can be obtained as 69.9%, 71.7% and 72.3% of recall, precision and Dice coefficient due to the increased sample size generated by the SMIG model. Therefore, the advantages of the SMIG model include privacy protection by generating synthetic data for research, medical data anonymising, which enables the future data sharing across multiple institutions and low-cost dataset generation procedure with high-quality MRIs. Nevertheless, there are few limitations of the paper. Firstly, the model has not been trained on other datasets for methodological validation. Secondly, the method has not integrated a mathematical model of glioblastoma growth in the TGP module.

The possibility of applying GAN for brain tumor growth prediction has been demonstrated in the reviewed papers. There are several directions for future works. Firstly, an extension of the current research can be the integration of multi-modal neuroimaging data, which should improve the prediction performance by leveraging rich information for glioblastoma from different medical imaging data sources. Secondly, the dataset collection for a larger sample size is of central significance for GAN-based methods. Data augmentation techniques and patch-based training strategy can also be considered to remedy the data shortage. Alternatively, the models should also be validated on animal experiments to evaluate the feasibility for human subjects in clinical scenarios.

## Discussion and Future Directions

GAN-based techniques have shown great promise in disease classification, anomaly and tumor detection, healthy brain aging modeling, Alzheimer's disease progression, and brain lesion evolution, as well as brain tumor growth modeling. Here, we briefly summarize the development of the adoption of GANs in neuroimaging and clinical neuroscience applications and provide future perspectives correspondingly.

**Disease classification.** GANs can identify patterns in high-dimensional, multi-modal imaging datasets and detect disease biomarkers at early stages. They can also enrich the dataset by synthesizing other modalities, and thereby boost classification performance through leveraging the complementary information provided by different modalities. In this section, we discussed the use of GANs in disease classification with a primary focus on Alzheimer's disease. We firstly examined GAN-based models on performing disease classification using single-modality data such as structural T1-weighted MRI and rs-fMRI.[55,57] Additionally, we investigated disease classification frameworks using multi-modal imaging data where GANs are used for missing data imputation such as PET images, which are usually scarcely available due to high cost and radioactive exposure, synthesis from MRI.[24,33,34,78,141] Functional measures are often studied in isolation but might provide important insights on behavior and cognitive ability, which are clinical markers of disease progression. In the future, we can potentially improve disease classification and diagnosis performance of neurodegenerative disease by leveraging both structural MRI and resting-state functional connectivity network components and relating the learned representations with cognitive measures.

**Tumor detection.** Tumor detection is a challenging task for machine learning systems due to the highly heterogeneous presentation of brain tumors as well as the relatively limited amount of labeled data available.[146] GANs have shown promise in tumor detection tasks, in both supervised and unsupervised anomaly detection.[23,80,83,84] In terms of dataset enrichment for supervised learning, GANs have difficulty modeling the heterogeneity of tumor data, given the limited sample size. On the other hand, GAN-based unsupervised anomaly detection only requires data from healthy subjects and is, instead, better suited to this problem. So far, many approaches have been proposed in GAN-based anomaly detection, but standardized benchmarks for principled evaluation and comparison are still absent.

**Brain aging.** In contrast to group/cohort analysis, where a one-size-fits-all approach is used to capture a homogenous pattern of brain aging for all subjects within a group,[91] GAN models reviewed in this section[88,93] help in capturing individual-level neuroanatomical variation which is conditioned on each subject's baseline brain anatomy, age, and diagnosis. Since many factors in addition to age and disease can influence brain changes,[136] a stratified approach for brain age modeling is necessary, which can further accommodate additive effects of confounding factors, such as genotype and lifestyle. However, GANs are often susceptible to mode collapse. Thus, learning individualized brain development is a non-trivial problem that requires careful model selection and training, as well as descriptive quantitative metrics to measure generative performance.

**Disease progression.** In this section, we discussed models that leverage GANs to simulate disease effects or stages using structural brain images and covariates, such as diagnosis and age.[36] To simplify the disease effects models, unrealistic assumptions are usually made in the literature. GAN has shown its ability in disentangling the structural heterogeneity of Alzheimer's disease in a baseline setting where assumptions are not necessary.[30] However, capturing disease effects in a longitudinal setting is non-trivial. Thus, extending these algorithms for longitudinal analysis is a potential promising future direction of modeling disease progression.

**Lesion progression.** We discussed two pioneering GAN-based studies in predicting the progression of white matter hyperintensities and multiple sclerosis separately in this section.[35,102] Longitudinal analysis is a challenging task especially at image level. Instead of directly estimating the dynamics of the brain lesions, authors from both works take the shortcut by leveraging GANs to generate the evolution maps from baseline scan to follow-up scan. These early attempts show the potential of GANs in tackling voxel-level lesion dynamics prediction which is clinically important for doctors' treatment decision making. Given the small sample size from both studies, GANs are very likely to be over-fitted to the training data as other deep learning methods in general. Thus, large-scale consortia need to be established to cover the diverse populations world-wide for unbiased precision medicine.

**Tumor growth.** Accurate prediction of brain tumor growth from non-invasive medical imaging scans is critical for disease prognosis and treatment planning. In this section, two GAN-based methods have demonstrated their ability in both predicting the glioma growth at a future time point and generating realistic tumor images for data augmentation.[110,111] In terms of longitudinal prediction, the study shows GANs empirically outperform other tumor growth models in the regime of limited sample size. In the data enrichment study, the tumor volume prediction task largely benefits from the increase of training dataset. These efforts illustrate the advantages of GANs in capturing the tumor growth characteristics by generating synthetic tumor scans. However, thorough validation of tumor growth prediction using GANs is necessary to evaluate the reliability of the model deployment at the clinical setting. Additionally, multimodal neuroimaging features might help in improving the growth prediction performance. Lastly, GANs-based tumor growth prediction algorithms can further facilitate survival prediction of the patients.

Besides these early successes, there are still limitations and challenges in developing diagnosis and longitudinal progression prediction models using these GAN-based methods. Here we discuss some potential issues and open questions to be addressed in the future.

**Cross-study robustness.** In order for deep learning models to be clinically useful, generalizability across multi-study datasets with diverse populations is important. GAN-based techniques perform well on single-site and single-study datasets, but previous research suggests that pooling data from multiple centers can reduce their statistical power and generalizability.[112,113,134] Data harmonization methods[114-119] offer a potential solution to tackle this challenge. For example, ComBat[114] is a linear mixed model that removes confounding site effects while preserving biological-relevant information in the data. Alternatively, models trained on a single-study can be adapted to unseen datasets with several strategies. The adversarial-robustness approach, for example, utilizes the discriminator from GAN to enforce the feature extractor to learn representations that are indistinguishable from the source study (model trained on) and the target study (unseen site). Similar to ComBat, GANs have also been used to explicitly "harmonize" scans across sites at image level.[120,121] In particular, unpaired image-to-image translation methods have demonstrated promise in mapping scans across sites while preserving anatomical and predictive signals.[122-124]

**Biological relevance.** GAN-based methods must generate biologically-relevant outputs for them to be useful. Current evaluation metrics[156,157] for GAN-based models primarily focus on measuring distances between real and generated images without verifying the preservation of basic underlying biological information. Specific metrics or evaluation procedures should be designed and utilized when we apply GAN-based models for synthesizing biomedical images. For example, one can compare the ability of both real and generated images to predict a variety of significant biological and clinical characteristics, including age, sex, cognitive performances, etc. Beyond post-hoc evaluation analyses, leveraging multi-modal data besides neuroimaging data during the training procedure has a better chance of generating biologically meaningful brain images. Given the growing evidence linking genetic variants and brain phenotypes,[158-160] it will be interesting to incorporate genetic data into model frameworks in order to generate imaging patterns or estimate disease effects underpinned by genetic differences or other biological mechanisms.

**Disease heterogeneity.** Neurological and neuropsychiatric diseases are often heterogeneous in neuroimaging and clinical phenotypes. GAN-based methods[30,125,126] have been applied to estimate various disease effects on neuroimaging features, and distill them into low-dimensional representations, which explain cognitive and clinical differences as well as possess discriminant and prognostic signals. However, the majority of these approaches cluster imaging data into discrete subtypes with variations contributed by both disease heterogeneity and severity. It is still challenging to efficiently model continuous disease progression processes and to parse disease severity and heterogeneity from imaging data independently.[126] Moreover, most of the recent methods only handle regional volume features derived from neuroimaging data. The GAN model can be potentially applied to voxel-wise data and learn heterogeneity from subtle imaging features. Lastly, existing methods were primarily trained with neuroimaging data only, while post hoc analyses were performed to further discover potentially associated genetic and clinical characteristics. A direct derivation of genetically driven disease subtypes or imaging signatures might provide more biological insights into disease heterogeneity and offer promise in precision medicine, targeted clinical trial recruitment, and drug discovery.

**Model interpretability.** The interpretability of machine learning methods is critical in explaining how pathological processes affect the structures of human brains.[135] However, nonlinear activation functions complicate the interpretation of deep learning-based models. Many algorithms have been proposed for addressing this problem, including saliency map,[132] class activation map,[133] etc. GAN-based methods offer another potential approach for model interpretation via data generation. For instance, in the context of modeling heterogeneous disease processes, GAN-based generative methods could explicitly synthesize a variety of patients' imaging data from healthy controls' data based on latent variables.[30,88,125,126] Comparisons of generated patient's data with input control's data, both visually and quantitatively, help in interpreting imaging patterns represented by the latent variables and dissect various neuroanatomical changes influenced by distinct underlying disease effects.

**Fairness-aware learning.** Unbalanced representation of demographics in the training data is another challenge when performing supervised/unsupervised learning using generative models. These models can amplify the bias in data and lead to undesirable performances towards underrepresented groups, such as females and African Americans.[127-129] This leads to concerns of health care equity and algorithmic fairness of ML-based diagnostic systems. Recently, several fairness-aware algorithms have been developed to tackle this challenge by using accessible demographic information or reweighting schemes for underrepresented samples. For example, FairGAN[130] utilizes a conditional GAN to generate balanced samples across demographic attributes and disease categories. Similarly, causal FairGAN[131] embeds a causal mechanism into the generator of a GAN to simulate complex data generating processes with multiple confounders.

**Multi-view learning.** Multi-modal MR images help in boosting the model prediction performance, but the incompleteness of the inputs, which is very common in practice, might be detrimental to the algorithm. Models should be robust to the situation when there are several absences of MR sequences from the patients to reduce the scanning expense.

**3D volume v.s. 2D image synthesis.** During the GAN design process, one important decision is whether to model the brain imaging data as a 3D volume or a sequence of 2D images or slices. The methodologies reviewed in this paper have used both 2D[88,102] as well as 3D[41,93] designs. The advantage of using 3D designs is that the GAN learns to synthesize whole brain volumes, thereby capturing 3-dimensional spatial features that can aid the disease classification or progression analysis. On the other hand, 2D designs are limited to capturing slice-level features that might not be sufficient for certain applications. For example, tumor growth modeling involves modeling volumetric changes of the tumor over time, hence a 3D design was adapted for this problem.[110] However, 3D GANs require large training data since the number of training parameters of the model explodes due to high dimensionality of the dataset. Moreover, computational and memory expenses increase significantly as the new dimension comes in. This might limit the adoption of 3D designs for certain applications where sufficient training data is not available. Another direct consequence of using 3D models is that they can be computationally expensive as they require optimization of larger neural networks.[137]

**Clinical deployment.** Translating machine learning models from the research stage to clinical practice for assisting decision making, is a critical step. However, these models, which are often developed under ideal conditions, face several challenges. Firstly, unlike research-oriented data that have passed quality check, clinical data are noisy and often incomplete, in terms of available sequences or modalities. Similar to other machine learning models, GAN-based methods should be robust to noisy input and be able to handle missing values. Secondly, distribution mismatch commonly happens between training (model development) and testing (model inference) data. We have discussed this challenge and potential solutions in the "cross-study robustness" section. Thirdly, deep learning models are usually developed with GPUs, which are not available in hospitals in general, to speed up the training and inference process. To achieve efficient model inference, cloud-based services, which have well-established machine learning infrastructure, are potential choices for hospitals to fill the gap of computing hardware and programming expertise.

Finally, the results produced by the GAN-based models should be easily interpretable for the doctors and practitioners. We have briefly discussed this concern in the "model interpretability" section.

**Complexity and reproducibility.** Both model and computational complexities of GAN-based methods are significantly greater than the traditional machine learning algorithms. For example, it can take from a few hours to a few days in GPU hours to train GANs as shown in Table 4 and 5. The high computational cost of GANs brings us the high-dimensional and realistic-looking synthesized images which are not possible by using traditional generative models. Although deep learning algorithms suffer from poor interpretability issues due to the non-linear mapping in general, GAN-based methods help in providing explainable visualizations on the model predictions where we discuss more in the "Model interpretability" section. Reproducibility is another important factor in evaluating the research work and their potential clinical use. Unfortunately, as shown in Table 4 and 5, only a few publications have shared their code and only one of them released the pre-trained model. The research community of machine learning in medical imaging should take the responsibility to release code and pre-trained parameters in order to encourage replicable research and accelerate the transition from research stage to clinical practice.

# Conclusions

In this review, we present the wide and successful adoption of a deep learning technique, namely generative adversarial networks, in neuroimaging and clinical neuroscience applications by briefly introducing the mechanism of GANs and showcasing their promises in several clinically meaningful tasks, including disease diagnosis, anomaly and tumor detection, brain development modeling, Alzheimer's progression estimation, lesion dynamics prediction, and tumor growth prediction. Based on the model architecture and experimental setup designs of each study covered in this review, we analyze the advantages and pitfalls of these algorithms from both technical soundness and clinical practice perspectives. In addition, given the gap between the current status of methodology development and clinical needs, we provide several timely future promising directions, such as algorithm reproducibility, interpretability, and fairness, which are critical in potential deployment of machine learning models.

# Acknowledgements

This work was supported by the National Institute on Aging (grant numbers RF1AG054409 and U01AG068057), the National Institute of Mental Health (grant number R01MH112070), the National Cancer Institute (grant number R01CA269948), the National Institute of Neurological Disorders and Stroke (grant number R01NS042645) and the National Institutes of Health (grant number 75N95019C00022).

# Data and code availability

No data or code was used for the research described in the article.

# Ethics statement

No human or animal study was involved in the article.

# Declaration of competing interest

The authors report no competing interest.

# Figures

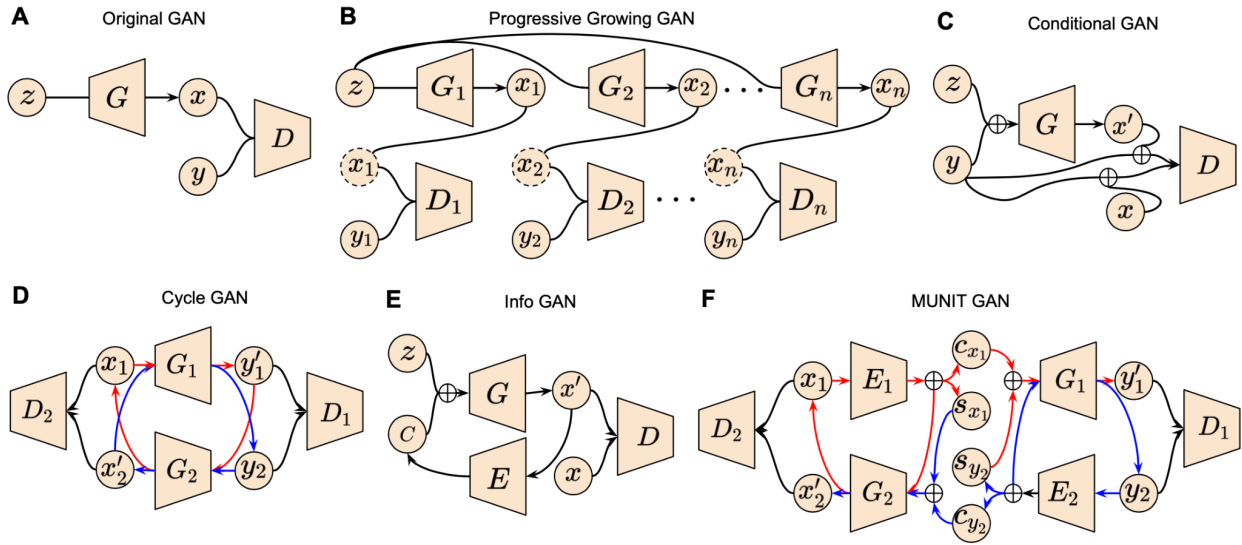

**Figure 1 Semantics of the original GAN and its extensions.** (**A**) Original GAN architecture where $z$, $x$ and $y$ denote random noise, generated image, and real image. $G$ and $D$ represent generator and discriminator separately. Wasserstein GAN, deep convolutional GAN, and self-attention GAN share the same structure of the original GAN but use a different loss function, convolutional block and self-attention module respectively. (**B**) Progressive growing GAN architecture. The resolution of each generated image $x_1$, $x_2$, ..., $x_n$ and real image $y_1$, $y_2$, ..., $y_n$ is increasing from left to right. The number of layers within each generator $G_1$, $G_2$, ..., $G_n$ and discriminator $D_1$, $D_2$, ..., $D_n$ is also growing accordingly. (**C**) Conditional GAN architecture where $z$, $y$, $x$, $x'$ denote random noise, extra label/information, real data, and generated data respectively. Concatenation of random noise $z$ and label $y$ are input to the generator and concatenation of label $y$ and generated/real data are input to the discriminator. (**D**) Cycle-GAN structure where $x_1$, $y_2$ denote umpired data from two different modalities, and $x_2'$, $y_1'$ denote generated data for the corresponding modality. $G_1$ transform data from modality $X$ to $Y$, and $G_2$ transforms inversely. Generated data $y_1'$ is reconstructed back to input $x_1$ through $G_2$

and same for generated data $x_2'$. (**E**) Info-GAN structure, where $z, c, x, x'$ denote random noise, informative part of latent variable, real data, and generated data. Concatenation of $z$ and $c$ are input to the generator. Informative latent $c$ are reconstructed through an encoder $E$ from generated data $x' = G(z, c)$. (**F**) MUNIT GAN structure where $x_1, y_2, x_2', y_1'$ denote unpaired data and generated data from two different modalities, $X$ and $Y$. $c_{x_1}, c_{y_2}, s_{x_1}, s_{y_2}$ denote content and style variables derived from data from two modalities respectively. Data $x_1$ is firstly decoded into content $c_{x_1}$ and style variable $s_{x_1}$ respectively. Then, the content variable $c_{x_1}$ is concatenated with a style variable from the $Y$ modality to generate data $y_1'$ through the generator $G_1$. Concatenation of $c_{x_1}$ and $s_{x_1}$ are used as input for reconstruct $x_1$ through $G_2$. Same process also applies to the reverse direction. Images are taken and adapted from.[12,28,31,37,38,40]

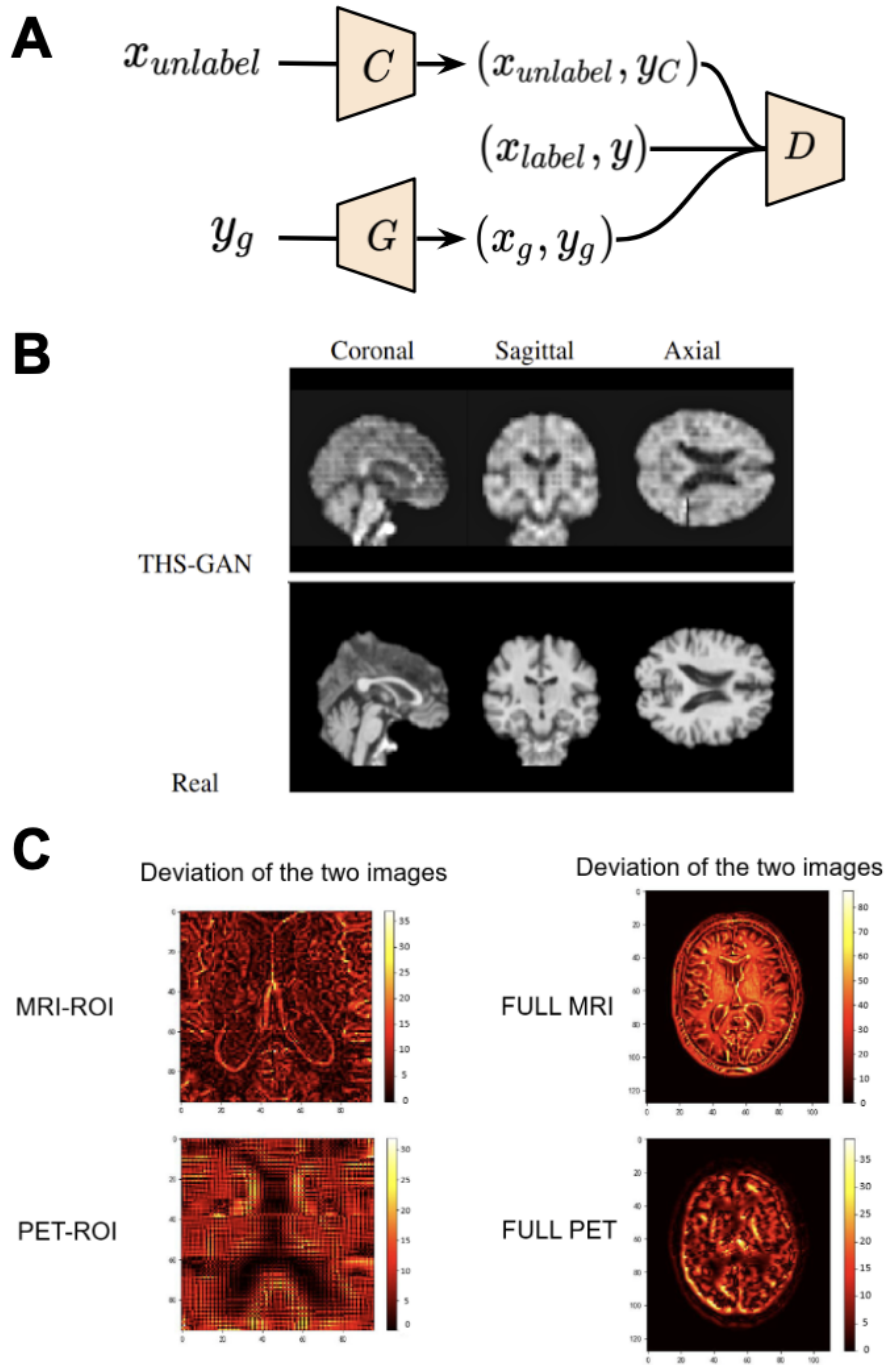

**Figure 2 GAN applications in disease classification with single- and multi-modal imaging.** (**A**) Schematic of THS-GAN for Alzheimer's disease and mild cognitive impairment classifications. (**B**) Comparison of synthesized brain MR images from THS-GAN and real T1-weighted scans with coronal, sagittal, and axial views for different training epochs. (**C**) Deviation between real image and synthetic images generated by Rev-GAN. In the deviation

image, the yellow color represents large differences, and the dark colors denote small deviations. Images are taken and adapted from.[33,41]

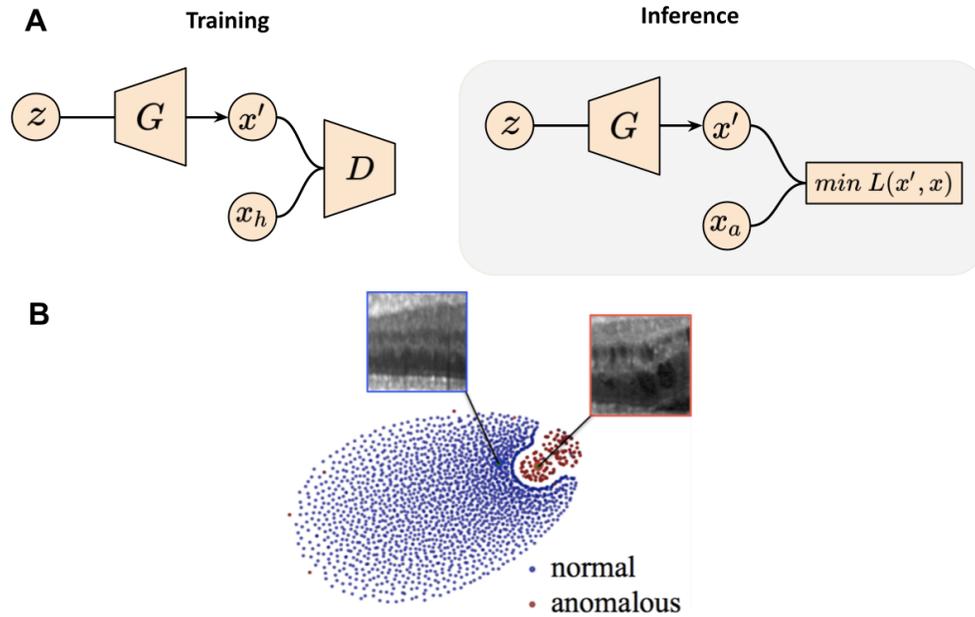

**Figure 3 GAN applications in neuroimaging-based anomaly detection.** (**A**) Schema of AnoGAN. Training is performed on health subjects to learn *z*, a latent space representing the data distribution. Inference is performed by sampling from *z* to generate image *x'*, such that the difference between *x'* and the anomalous image $x_a$ is minimized. (**B**) Illustration of the latent space distribution produced by AnoGAN. Images are taken and adapted from.[83]

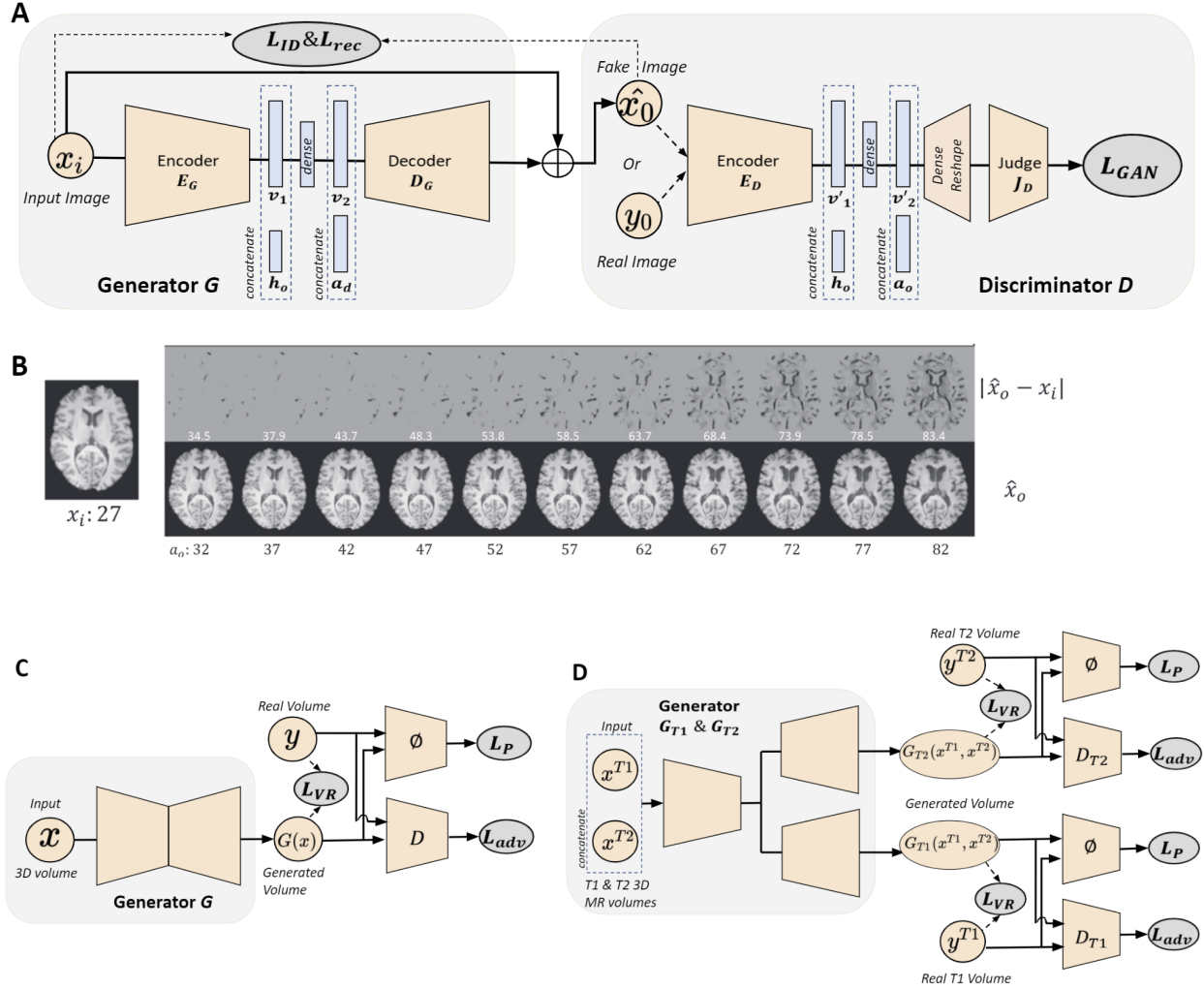

**Figure 4 GAN applications in modeling healthy brain aging.** (**A**) Schematic of the conditional GAN model for modeling the brain aging process across the whole lifespan. $x_i$: generator input; $h_0$: target age vector; $a_d$: age difference between current age $a_i$ and target age $a_0$; $\hat{x}_0$: generator output; $v_1$, $v_2$, $v'_1$, $v'_2$: latent embedding. Generator synthesizes brain image of target age and health state, and judge network gives a discrimination score of whether the image given to the discriminator is real or fake. $L_{ID}$, $L_{rec}$, $L_{GAN}$ refer to identity loss, reconstruction loss and adversarial loss respectively. (**B**) Examples of healthy brain aging modeling using the GAN described in (A). Bottom panel shows the images synthesized at different target ages $a_0$, and the top panel shows the absolute difference between input image $x_i$ and synthesized image $\hat{x}_0$. (**C**) Schematic of the perceptual adversarial network (PGAN). (**D**) Multi-modal perceptual

adversarial network (MPGAN) architecture. $x, x^{T1}, x^{T2}$: input 3D MR volume; $G(x), G_{T1}(x^{T1}, x^{T2}), G_{T2}(x^{T1}, x^{T2})$: generated output; $y, y^{T1}, y^{T2}$: real 3D MR volume; $D, D_{T1}, D_{T2}$: discriminator networks; ϕ: feature extraction network; $L_{VR}, L_{P}, L_{adv}$ refer to voxel-wise reconstruction loss, perceptual loss, and adversarial loss. Images are taken and adapted from.[88,93]

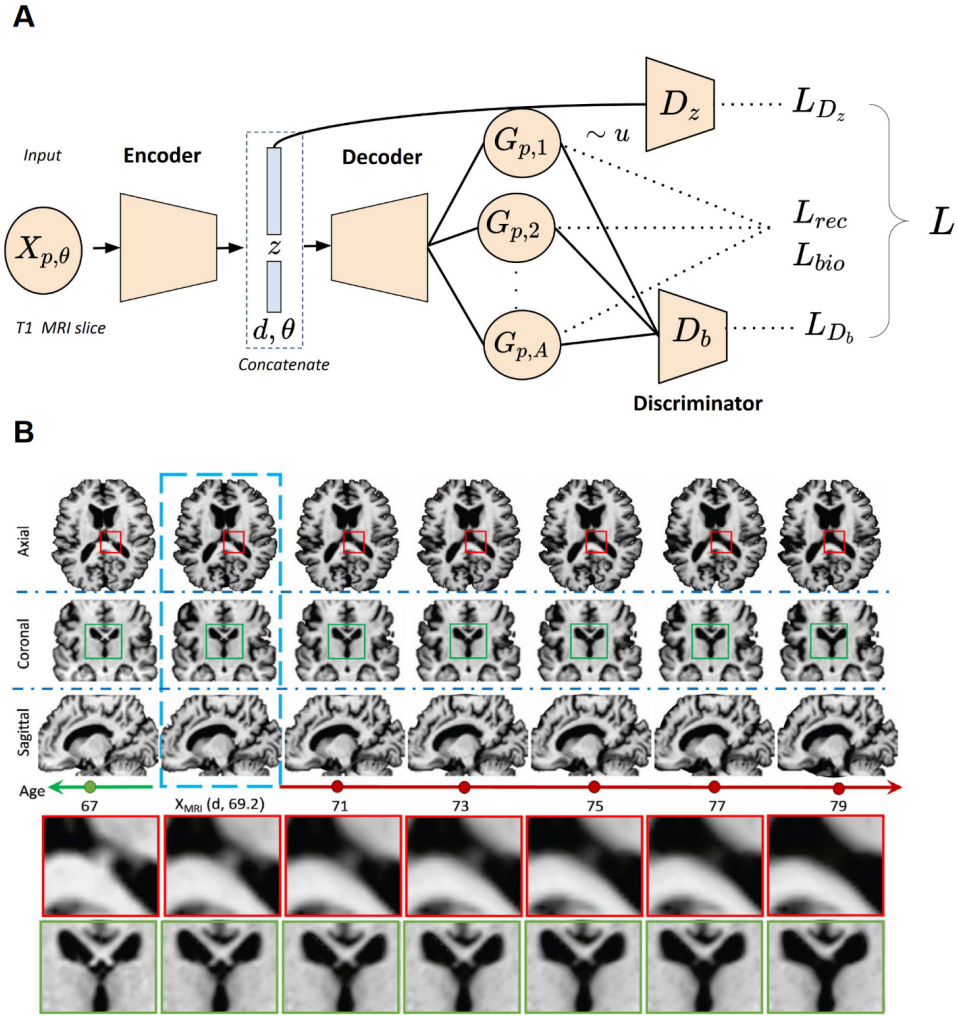

**Figure 5 GAN applications in generating disease progression scans from a single time point.** (**A**) Schematic of DANI net. The input to DANI net is a T1 MR image from subject $p$ at age $\theta$ with diagnosis $d$. The output of the decoder is a set of longitudinal scans. Several loss functions (reconstruction loss $L_{rec}$, biological constraints $L_{bio}$, discriminator losses $L_{D_z}$ and $L_{D_b}$) are combined together to train DANI net using a single time point of subject $p$. (**B**) Longitudinal MRIs synthesized using 4D-DANI-Net for a 69 years old cognitive normal subject from three orientations. The blue box indicates the input MRI and other images are synthesized MR scans from the model. Two magnified regions are illustrated at the bottom panel. Images are taken and adapted from.[96]

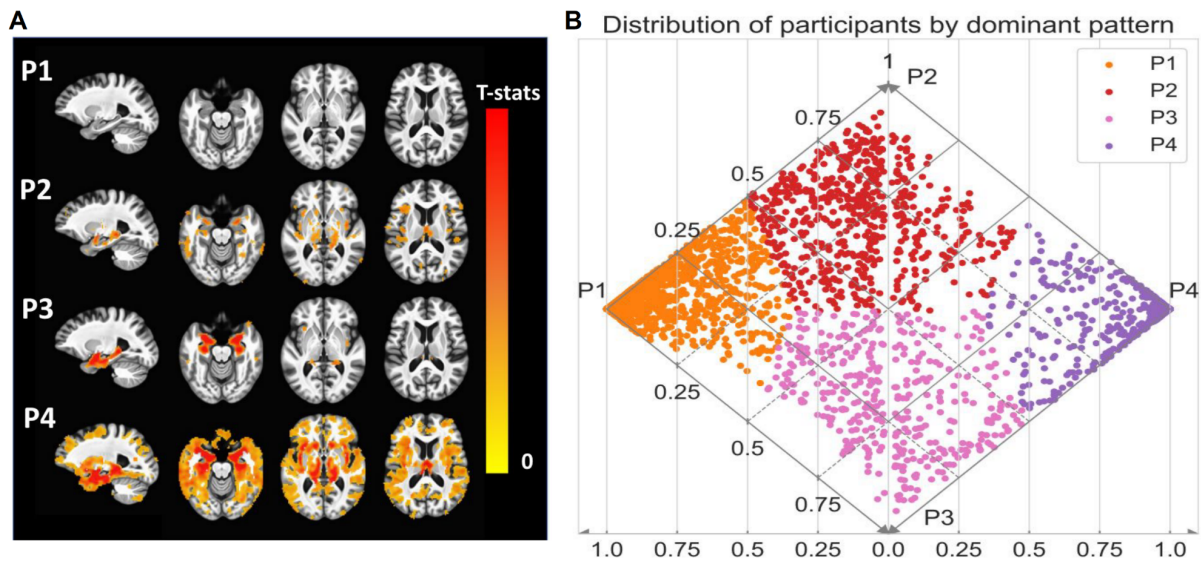

**Figure 6 GAN applications in disease subtypes discovery (four-dimensional coordinate system developed by SMILE-GAN).** (**A**) Voxel-wise statistical comparison (one-sided t-test) between cognitive normal subjects and subjects that predominantly belong to each of the four Alzheimer's disease neuroanatomical patterns. (**B**) Visualization of subjects that belong to the four subtype clusters in a diamond plot. Images are taken and adapted from.[30]

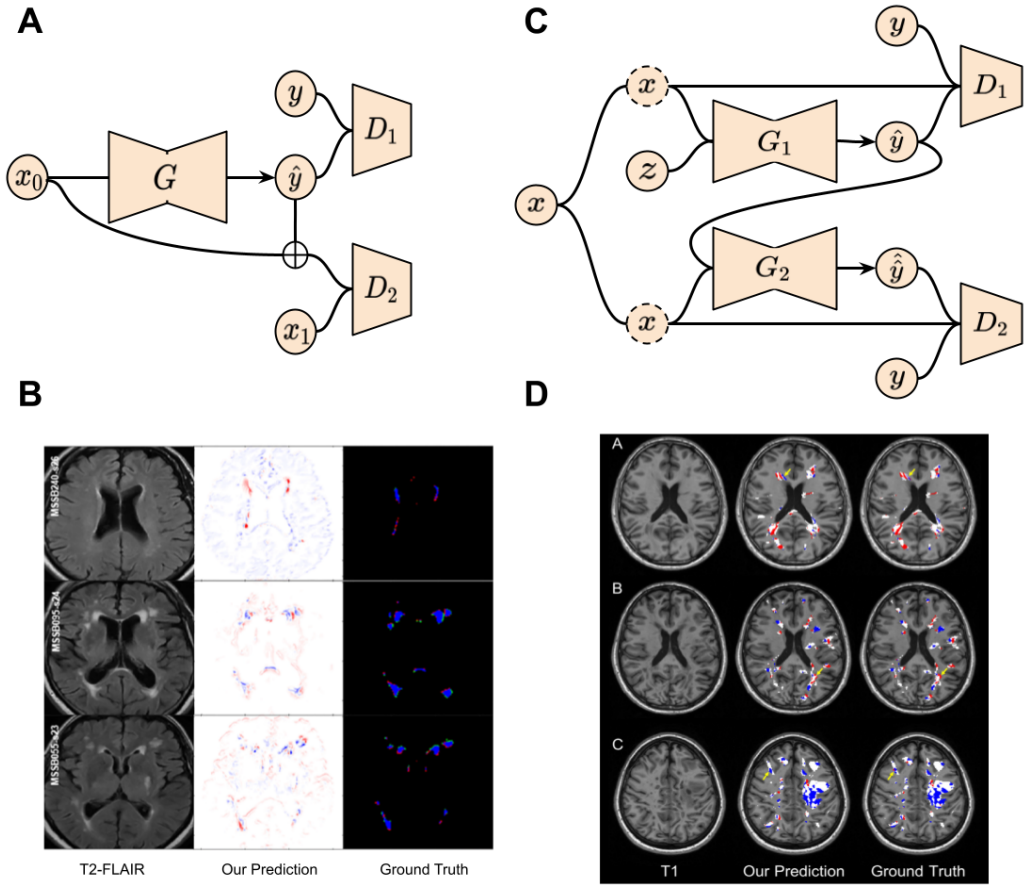

**Figure 7 GAN applications in brain lesion (white matter hyperintensities and multiple sclerosis) evolution prediction.** (**A**) Schematic of the GAN for white matter hyperintensities evolution prediction. (**B**) Disease evolution map examples produced by GAN and the derived irregularity map from two time points. (**C**) Two-stage conditional GAN for [$^{11}$C] PIB PET images generation from multi-sequence MR images for myelin content in multiple sclerosis dynamic prediction. (**D**) Examples of myelin content changes indicating demyelination (red color) and remyelination (blue color) from both GAN outputs and real PET images. Images are taken and adapted from.[35,102]

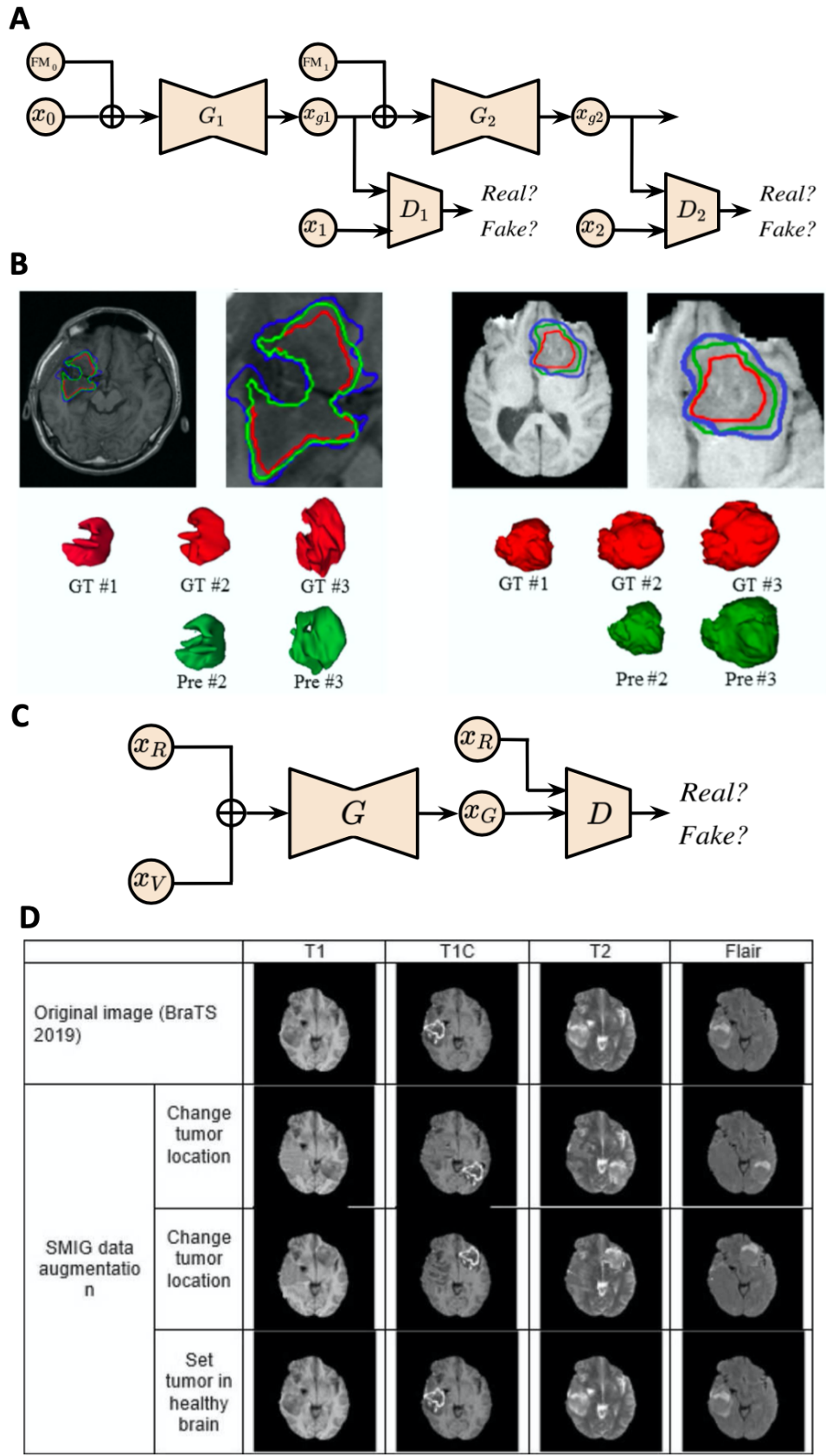

**Figure 8 GAN applications in brain tumor growth prediction.** (A) GP-GAN architecture for glioma growth prediction. $x_{gi}$: generated image at time point $i$; $G_i$: generator at time point $i$; $D_i$: discriminator at time point $i$. (**B**) Growth prediction for subjects with low-grade glioma (left) and high-grade glioma (right) at different time points via GP-GAN. GT: ground truth; Pre: prediction. (**C**) Schematic of SMIG model. The model is trained to 1) generate an abnormal brain based on a healthy brain from ADNI dataset and tumor volume from TCIA; 2) change tumor location. $x_R$: image represents a healthy brain or tumor in real location; $x_V$: tumor volume provided by TCIA; $x_G$: generated image; $G$: generator; $D$: discriminator. (**D**) SMIG model applications on single patient images from BraTS dataset. Images are taken and adapted from.[110,111]

# Tables

**Table 1 Frequently applied GAN architectures in neuroimaging.** Publications are ordered by year in ascending order.

| Publication | Highlights |
| --- | --- |
| GAN (Goodfellow et al., 2014) | Original GAN |
| CGAN (Mirza and Osindero, 2014) | Conditional GAN |
| InfoGAN (Chen et al., 2016) | Interpretable representation learning |
| CycleGAN (Zhu et al., 2017) | Unpaired image-to-image translation |
| WGAN (Gulrajani et al., 2017) | Wasserstein GAN |
| PGGAN (Karras et al., 2018) | Progressive growing GAN |
| MUNIT (Huang et al., 2018) | Multi-modal unsupervised image-to-image translation |
| SAGAN (Zhang et al., 2019a) | Self-attention GAN |
| ClusterGAN (Mukherjee et al., 2019) | Clustering GAN |
| Rev-GAN (van der Ouderaa and Worrall, 2019) | Reversible GAN |
| StyleGAN (Karras et al., 2019) | Style-transfer GAN |

**Table 2 Overview of publications utilizing GANs to assist disease diagnosis.** Publications are clustered by categories and ordered by year in ascending order.

| Publication | Method | Dataset | Modality | Highlights |
| --- | --- | --- | --- | --- |
| **Disease Classification** | | | | |
| (Yan et al., 2018) | Conditional GAN | ADNI | MRI, PET | Amyloid PET generation and MCI prediction |
| (Liu et al., 2020) | GAN | ADNI | MRI, PET | MCI conversion prediction |
| (Mirakhorli et al., 2020) | GNN & GAN | ADNI | rs-fMRI | AD-related patterns extraction |
| (Pan et al., 2021b) | Feature-consistent GAN | ADNI | MRI, PET | Joint synthesis and diagnosis |
| (Lin et al., 2021) | Reversible GAN | ADNI | MRI, PET | Bidirectional mapping between modalities |
| (Gao et al., 2021) | Pyramid and Attention GAN | ADNI | MRI, PET | Missing modality imputation |
| (Yu et al., 2021) | Higher-order pooling GAN | ADNI | MRI | Semi-supervised learning |
| (Pan et al., 2021a) | Decoupling GAN | ADNI | DTI, rs-fMRI | Abnormal neural circuits detection |
| (Zuo et al., 2021) | Hypergraph perceptual network | ADNI | MRI, DTI, rs-fMRI | Abnormal brain connections analysis |
| **Tumor Detection** | | | | |
| (Han et al., 2019) | PGGAN & MUNIT | BRATS | MRI | Tumor image augmentation |
| (Huang et al., 2019) | Context-aware GAN | BRATS | T1, T2, FLAIR | Glioma severity grading |
| (Park et al., 2021) | StyleGAN | Private | T1, T2, FLAIR | IDH-mutant glioblastomas generation |
| **Anomaly Detection** | | | | |
| (Wei et al., 2019) | Sketcher-Refiner GAN | Private | PET, DTI | Myelin content in Multiple Sclerosis |

**Table 3 Overview of publications utilizing GANs to assist brain development and disease progression analysis.** Publications are clustered by categories and ordered by year in ascending order.

| Publication | Method | Dataset | Modality | Highlights |
| --- | --- | --- | --- | --- |
| **Brain Aging** | | | | |
| Xia et al. (2021) | Transformer GAN | Cam-CAN, ADNI | MRI | Synthesize aging brain without longitudinal data |
| Peng et al. (2021) | Perceptual GAN | IBIS | T1, T2 | Infant brain longitudinal imputation |
| **Alzheimer's Progression** | | | | |
| Bowles et al. (2018) | WGAN | ADNI | MRI | Disentangle visual appearance of AD using latent encoding |
| Wegmayr et al. (2019) | Recursive GAN | ADNI, AIBL | MRI | Conversion prognosis from MCI to AD |
| Zhao et al. (2020b) | Multi-information GAN | ADNI, OASIS | MRI | Progression Stage classification |
| Yang et al. (2021) | Cluster & Info-GAN | ADNI, BLSA | MRI | AD subtypes imaging patterns discovery |
| Ravi et al. (2022) | Conditional GAN | ADNI | MRI | Spatiotemporal, biologically-informed constrained |
| **Lesion Evolution** | | | | |
| Rachmadi et al. (2020) | Multi-discriminator GAN | Private | T1, T2, FLAIR | White matter hyperintensities evolution prediction |
| Wei et al. (2020) | Conditional Attention GAN | Private | MRI, PET | Myelin content prediction in multiple sclerosis |
| **Tumor Growth** | | | | |
| Elazab et al. (2020) | Stacked Conditional GAN | Private, BRATS | T1, T2, FLAIR | Glioma growth prediction |
| Kamli et al. (2020) | GAN | TCIA, ADNI | T1, T2, FLAIR | Glioblastoma tumors growth prediction |

**Table 4 Computational complexity and reproducibility of disease diagnosis-specific GANs publications.** Publications are clustered by categories and ordered by year in ascending order.

| Publication | Method | GPU Time (hr) | Code | Pre-trained Model |
| --- | --- | --- | --- | --- |
| **Disease Classification** | | | | |
| (Yan et al., 2018) | Conditional GAN | N/A | N/A | N/A |
| (Liu et al., 2020) | GAN | N/A | N/A | N/A |
| (Mirakhorli et al., 2020) | GNN & GAN | N/A | N/A | N/A |
| (Pan et al., 2021b) | Feature-consistent GAN | N/A | N/A | N/A |
| (Lin et al., 2021) | Reversible GAN | N/A | N/A | N/A |
| (Gao et al., 2021) | Pyramid and Attention GAN | N/A | N/A | N/A |
| (Yu et al., 2021) | Higher-order pooling GAN | N/A | N/A | N/A |
| (Pan et al., 2021a) | Decoupling GAN | N/A | N/A | N/A |
| (Zuo et al., 2021) | Hypergraph perceptual network | 8 | N/A | N/A |
| **Tumor Detection** | | | | |
| (Han et al., 2019) | PGGAN & MUNIT | N/A | N/A | N/A |
| (Huang et al., 2019) | Context-aware GAN | N/A | N/A | N/A |
| (Park et al., 2021) | StyleGAN | N/A | N/A | N/A |
| **Anomaly Detection** | | | | |
| (Wei et al., 2019) | Sketcher-Refiner GAN | N/A | N/A | N/A |

**Table 5 Computational complexity and reproducibility of brain development and disease progression analysis-specific GANs publications.** Publications are clustered by categories and ordered by year in ascending order.

| Publication | Method | GPU Time (hr) | Code | Pre-trained Model |
| --- | --- | --- | --- | --- |
| **Brain Aging** | | | | |
| Xia et al. (2021) | Transformer GAN | N/A | https://github.com/xiat0616/BrainAgeing | N/A |
| Peng et al. (2021) | Perceptual GAN | N/A | https://github.com/liying-peng/MPGAN | N/A |
| **Alzheimer's Progression** | | | | |
| Bowles et al. (2018) | WGAN | N/A | N/A | N/A |
| Wegmayr et al. (2019) | Recursive GAN | 144 | https://github.com/vwegmayr/brain-aging | N/A |
| Zhao et al. (2020b) | Multi-information GAN | N/A | N/A | N/A |
| Yang et al. (2021) | Cluster & Info-GAN | N/A | https://github.com/zhijian-yang/SmileGAN | N/A |
| Ravi et al. (2022) | Conditional GAN | 72 | https://github.com/daniravi/Brain-MRI-Simulator | N/A |
| **Lesion Evolution** | | | | |
| Rachmadi et al. (2020) | Multi-discriminator GAN | N/A | https://github.com/febrianrachmadi/dep-gan-im | https://github.com/febrianrachmadi/dep-gan-im |
| Wei et al. (2020) | Conditional Attention GAN | N/A | N/A | N/A |
| **Tumor Growth** | | | | |
| Elazab et al. (2020) | Stacked Conditional GAN | N/A | N/A | N/A |
| Kamli et al. (2020) | GAN | N/A | N/A | N/A |